\pdfoutput=1
\documentclass[11pt]{article}
\usepackage[]{ACL2023}

\usepackage{times}
\usepackage{latexsym}
\usepackage{hyperref}
\usepackage{color, colortbl}
\definecolor{Gray}{gray}{0.9}
\usepackage{array}
\usepackage{booktabs}
\usepackage{multirow}
\usepackage{amsmath}
\usepackage{amssymb}
\usepackage{graphicx}
\usepackage{enumerate}
\usepackage{diagbox}
\usepackage{mathtools}
\usepackage{paralist}
\usepackage{tabularx}
\usepackage{lipsum}
\usepackage{ragged2e}

\usepackage[T1]{fontenc}

\usepackage[utf8]{inputenc}

\usepackage{microtype}

\usepackage{inconsolata}

\newcommand{\wq}[1]{\textcolor{black}{#1}}

\newcommand*{\affaddr}[1]{#1} 
\newcommand*{\affmark}[1][*]
{\textsuperscript{#1}}

\title{CAT: A Contextualized Conceptualization and Instantiation Framework for Commonsense Reasoning}

\author{
Weiqi Wang\affmark[1]\thanks{\quad Equal Contribution}~, 
Tianqing Fang\affmark[1]$^{*}$,
Baixuan Xu\affmark[1],
Chun Yi Louis Bo\affmark[1],\\
\textbf{Yangqiu Song\affmark[1],
Lei Chen\affmark[1,2]}\\
\affaddr{\affmark[1]Department of Computer Science and Engineering, HKUST, Hong Kong SAR, China}\\
\affaddr{\affmark[2]Information Hub, HKUST (GZ), Guangzhou, China} \\
\texttt{\{wwangbw, tfangaa\}@cse.ust.hk, bxuan@connect.ust.hk} \\ 
\texttt{\{cybo, yqsong\}@cse.ust.hk, leichen@ust.hk}
}

\begin{document}
\maketitle
\begin{abstract}
Commonsense reasoning, aiming at endowing machines with a human-like ability to make situational presumptions, is extremely challenging to generalize.
For someone who barely knows about \textit{meditation}, while is knowledgeable about \textit{singing}, he can still infer that \textit{meditation makes people relaxed} from the existing knowledge that \textit{singing makes people relaxed} by first conceptualizing \textit{singing} as a \textit{relaxing event} and then instantiating that event to \textit{meditation}.  
This process, known as conceptual induction and deduction, is fundamental to commonsense reasoning while lacking both labeled data and methodologies to enhance commonsense modeling.
To fill such a research gap, we propose CAT (\textbf{C}ontextualized Conceptu\textbf{A}lization and Ins\textbf{T}antiation),
a semi-supervised learning framework that integrates event conceptualization and instantiation to conceptualize commonsense knowledge bases at scale.
Extensive experiments show that our framework achieves state-of-the-art performances on two conceptualization tasks, and the acquired abstract commonsense knowledge can significantly improve commonsense inference modeling.
Our code, data, and fine-tuned models are publicly available at~\href{https://github.com/HKUST-KnowComp/CAT}{https://github.com/HKUST-KnowComp/CAT}.
\end{abstract}

\section{Introduction}

\begin{figure}[h]
    \centering
    \includegraphics[width=1\linewidth]{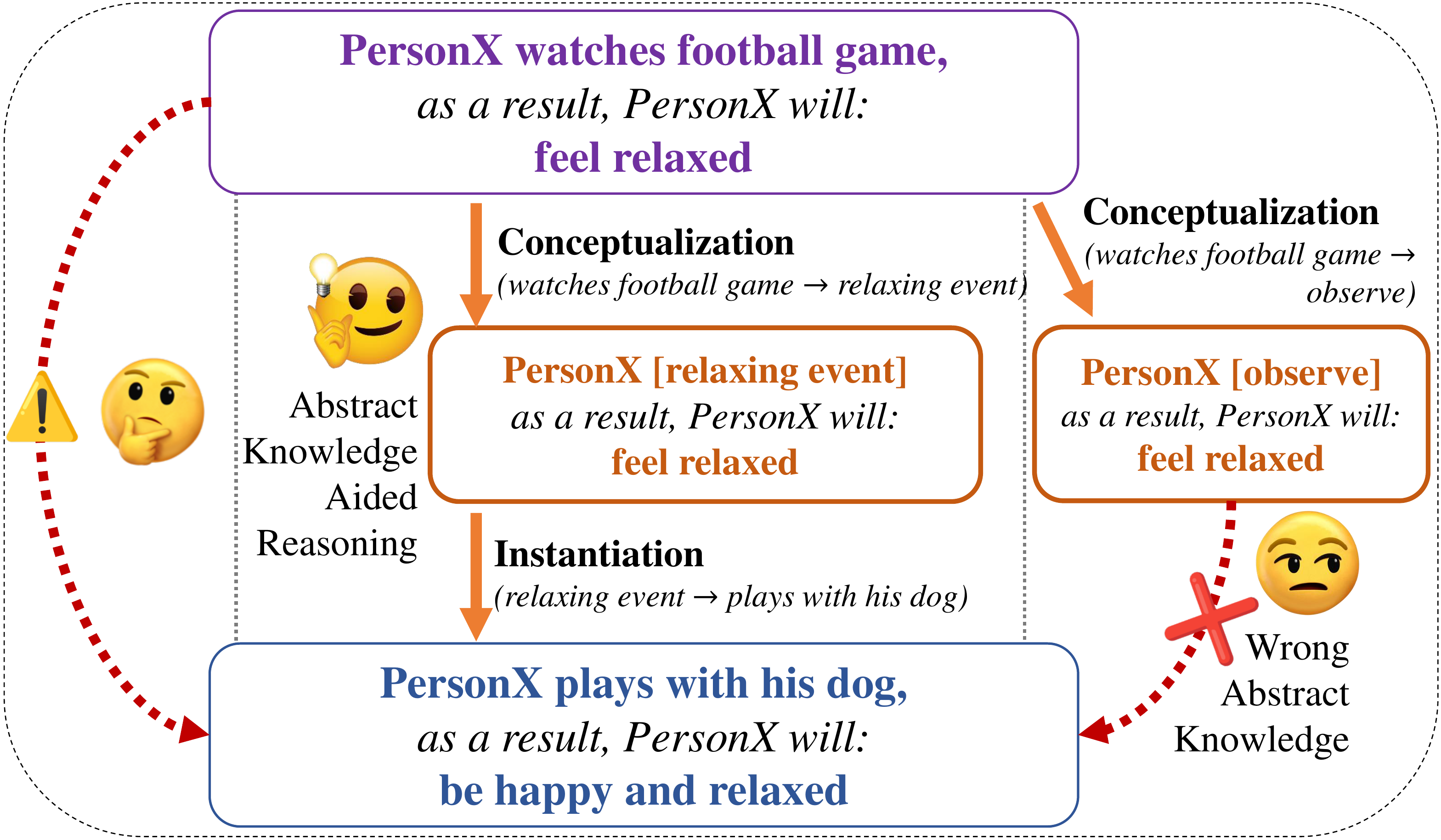}
    \caption{A demonstration of commonsense reasoning on an unknown situation, \textit{PersonX plays with his dog}, with the aid of {\textcolor{orange}{abstract commonsense knowledge}}. 
    Decontextualized conceptualization, such as \textit{observe}, may yield wrong {\textcolor{orange}{abstract commonsense knowledge}} that cannot be instantiated within the corresponding context.}
    \label{fig:Introduction_demo}
    \vspace{-0.1in}
\end{figure}

\begin{quote}
  ``\textit{Concepts are the glue that holds our mental world together.}''-- \citet{murphy2004big}
\end{quote}

Commonsense reasoning is a crucial ability for machines to make situational presumptions and draw inferences from the knowledge that reflects our humans' understanding of situations and common facts~\cite{DBLP:books/daglib/0066824,DBLP:journals/cacm/DavisM15}.
It has gained increasing popularity in the Natural Language Processing (NLP) community with the emergence of CommonSense Knowledge Bases (CSKB)~\cite{DBLP:conf/aaai/SapBABLRRSC19,DBLP:conf/aaai/SpeerCH17,DBLP:conf/aaai/HwangBBDSBC21} and large language models~\cite{DBLP:conf/acl/BosselutRSMCC19,DBLP:conf/acl/RajaniMXS19,DBLP:conf/acl/0010LLWWBCH22,su-etal-2022-mico, DBLP:conf/acl/YuZSN22}.
However, when encountering situations beyond the data given, more abstract background knowledge must be acquired and generalized to assist the reasoning~\cite{tenenbaum2011grow}, and language models trained with an autoregressive language modeling objective do not explicitly leverage such abstract knowledge during inference.

Instead, humans rely on conceptual induction and deduction~\cite{murphy2004big} to make inferences on novel situations without the need to memorize all special cases.
As shown in Figure~\ref{fig:Introduction_demo},
humans can derive conceptualizations based on the assertion that ``PersonX watches a football game, as a result, he feels relaxed'' to infer that ``relaxing events can make someone feel relaxed,''
where the acquired abstract commonsense knowledge can be further used as general knowledge to perform reasoning on similar or associated situations.
A new commonsense knowledge ``PersonX plays with his dog, as a result, he feels happy and relaxed'' can be deduced by instantiating \textit{relaxing events} to \textit{playing with his dog}.
As the cornerstone of generalizable commonsense reasoning, such a process is extremely challenging for machines to replicate due to the absence of contextualized conceptualizations and abstract commonsense knowledge in CSKBs and a lack of relevant methodologies.

Yet, existing works address the process of induction and deduction separately via conceptualization and instantiation.
Several methods performing conceptualization are proposed with a specific focus on entity-level~\cite{DBLP:conf/eacl/DurmeMS09,DBLP:conf/ijcai/SongWWLC11,DBLP:conf/aaai/GongZZ16,DBLP:journals/corr/abs-2003-03239,DBLP:journals/corr/abs-2211-04079,DBLP:conf/ijcai/SongWW15} and event-level~\cite{DBLP:conf/conll/ChenZWR20,DBLP:journals/corr/abs-2206-01532} semantics.
Instantiation~\cite{DBLP:journals/corr/abs-2205-11658}, as the process that simulates conceptual deduction, is tackled separately and not leveraged by these methods.
Though abstract commonsense knowledge can be derived by using existing conceptualization methods to abstract a certain instance from factual commonsense knowledge, several limitations still exist.

First, the plausibility of abstract commonsense knowledge banks on both the correctness of \textit{conceptualization} and proper \textit{contextualization} under specific assertions.
The latter one, which is an essential step for the deduction of abstract knowledge, is missing from current methodologies.
Take Figure~\ref{fig:Introduction_demo} as an example, the concept \textit{observe} will not necessarily lead to the result of ``feeling relaxed'', as \textit{observe} omits the entertaining property of the original instance as a cost of abstraction. 
Second, instantiating abstract commonsense knowledge can yield much more and diverse concrete commonsense knowledge that can serve as an augmentation of the training dataset, while current methods undervalue such a process and only focus on conceptualization.
Finally, the complex \textit{contextualization} and \textit{conceptualization} of commonsense knowledge 
can easily bring more than two orders of magnitude of data on top of the original dataset.
This makes current labeled data scarce and infeasible for practitioners to annotate all of them, leaving a large amount of unlabeled data.

To fill in these research gaps, we propose CAT (\textbf{C}ontextualized Conceptu\textbf{A}lization and Ins\textbf{T}antiation), a semi-supervised learning framework that unites event conceptualization and instantiation in cascade to conceptualize CSKBs and acquire abstract commonsense knowledge to aid commonsense reasoning.
Inspired by how humans learn with concepts~\cite{carey2004bootstrapping}, 
we design a novel bootstrapping\footnote{Bootstrapping refers to the linguistics term in language acquisition that humans learn new knowledge by recognizing its semantic elements and connecting them with known knowledge~\cite{pinker1987bootstrapping}.}
method to enhance conceptualizations and abstract commonsense knowledge verification with the help of similar conceptualizations and instantiations as a reference.
We demonstrate the effectiveness of CAT by using the acquired abstract commonsense knowledge to train COMET~\cite{DBLP:conf/acl/BosselutRSMCC19}, a commonsense inference language model that generates if-then commonsense knowledge, and showing that our derived abstract commonsense knowledge can significantly improve commonsense inference modeling.

Our contributions are three-fold:
(1) We introduce a semi-supervised learning framework, CAT, to conceptualize CSKBs with the assistance of progressively bootstrapping similar abstract concepts or instantiations in the conceptualization process.
(2) We use CAT to acquire abstract commonsense knowledge at scale with high quality, which can be used for commonsense inference modeling. 
(3) We demonstrate the effectiveness of our framework by achieving state-of-the-art performance on two CSKB conceptualization tasks and remarkably improving commonsense inference modeling with our derived abstract commonsense knowledge.

\section{Related Works}
\paragraph{Conceptualization and Instantiation.}

Many existing works have studied conceptualization and instantiation separately.
\citet{DBLP:conf/eacl/DurmeMS09} first attempted to derive more general knowledge by abstracting over large sets of factoids obtained from WordNet~\cite{DBLP:journals/cacm/Miller95} synsets.
\citet{DBLP:conf/ijcai/SongWWLC11,DBLP:conf/ijcai/SongWW15} and \citet{DBLP:conf/aaai/GongZZ16} proposed to turn instances in a sentence into concepts via weight matching from Probase~\cite{DBLP:conf/sigmod/WuLWZ12}.
Recently, \citet{DBLP:journals/ai/LiuCWLCXCJ22} proposed a taxonomy-guided induction method to mine verb-oriented commonsense knowledge from verb phrases.
\citet{DBLP:journals/corr/abs-2211-04079} constructed a conceptual knowledge benchmark to evaluate language models with three zero-shot probing tasks.
While these works focus on the conceptualization of entities, \citet{DBLP:journals/corr/abs-2206-01532} constructed an event conceptualization benchmark based on ATOMIC~\cite{DBLP:conf/aaai/SapBABLRRSC19} by combining syntactic parsing, semantically heuristic matching, and human annotation.
Besides, the line of works focusing on ultra-fine entity typing~\cite{DBLP:conf/acl/LevyZCC18,DBLP:conf/acl/DaiSW20,DBLP:journals/tacl/LiYC22} shared similar objectives of typing named entities, nominal nouns, and pronouns into a set of free-form phrases.
Instantiation was attempted by~\citet{DBLP:journals/corr/abs-2205-11658}, who proposed a controllable generative framework to probe valid instantiations for abstract knowledge automatically.
Though~\citet{DBLP:conf/naacl/PoradaSTC21} and~\citet{DBLP:journals/corr/abs-2211-04079} both proved that existing pretrained language models lack conceptual knowledge, none of existing works explicitly combine both techniques to derive abstract knowledge that is context-sensitive and generalizable.

\paragraph{Commonsense Reasoning.}

Endowing NLP systems with the ability to perform commonsense reasoning is an elusive goal of artificial intelligence~\cite{DBLP:conf/acl/SapSBCR20}.
A diverse collection of commonsense reasoning tasks have been proposed as evaluation benchmarks~\cite{DBLP:conf/naacl/TalmorHLB19,DBLP:conf/emnlp/OmuraKK20,DBLP:conf/emnlp/PontiGMLVK20,DBLP:conf/emnlp/FangWCHZSH21}.
Among them, \citet{DBLP:conf/acl/BosselutRSMCC19} proposed a generative model, COMET, to learn to produce \textit{if-then} commonsense knowledge as an effective approach toward modeling commonsense inference that can be applied in various commonsense reasoning tasks~\cite{DBLP:conf/naacl/TalmorHLB19}.

\paragraph{Semi-Supervised Learning.}
Semi-supervised learning (SSL) aims at taking advantage of unlabeled data to equip models with stronger generalization ability~\cite{DBLP:journals/ml/EngelenH20}.
The most common approach is using pseudo labels~\cite{DBLP:conf/cvpr/IscenTAC19,DBLP:conf/naacl/WangT0G22} to expose more unseen data to the student model.
It has been applied in various machine learning tasks such as image classification~\cite{DBLP:conf/cvpr/LiuTCLBC22a,DBLP:conf/cvpr/HuYHN21}, text classification~\cite{DBLP:conf/emnlp/LiZC0SY21,DBLP:conf/aaai/MengSZH19, DBLP:conf/www/XiaoLS19}, commonsense knowledge base population~\cite{DBLP:journals/corr/abs-2210-07988}, and named entity recognition~\cite{DBLP:conf/naacl/LiuFTCZHG21,DBLP:conf/acl/ChenJW0G20}.

\begin{figure*}[t]
    \centering
    \includegraphics[width=1\linewidth]{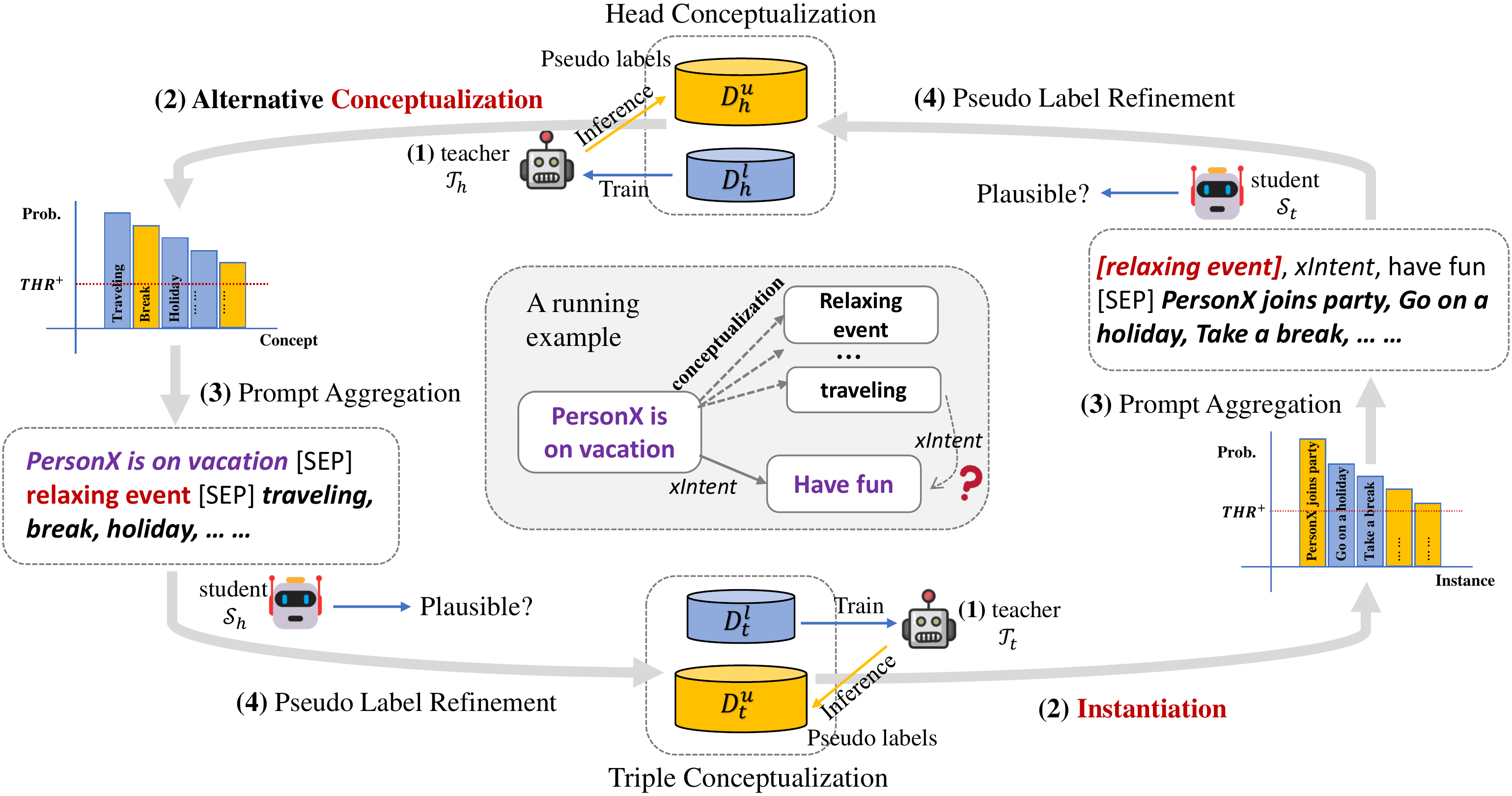}
    \caption{Overview of our CAT framework. 
    A running example that conceptualizes the triple (PersonX is on vacation, \texttt{xIntent}, have fun) is presented in the figure, where the head is conceptualized first, and the model needs to determine whether the conceptualized triple still holds after the event conceptualization.}
    \label{fig:CAT_overview}
\end{figure*}

\begin{table}[t]
\centering
\small
\begin{tabular}{@{}l|l|ccc@{}}
\toprule
Data & Type & Train & Dev & Test \\ 
\midrule
\multirow{2}{*}{$D^l$} & \#event & 107,384 & 12,117 & 11,503 \\
 & \#triple & 65,386 & 8,403 & 7,408 \\ 
 \midrule
\multirow{2}{*}{$D^u$} & \#event & 304,983 & 36,023 & 31,578 \\
 & \#triple & 4,851,272 & 499,523 & 570,400 \\ 
 \bottomrule
\end{tabular}
\vspace{-0.5em}
\caption{Statistics of labeled data $D^l$ and unlabeled data $D^u$ in AbstractATOMIC.}
\label{tab:AbstractAtomic_statistics}
\vspace{-0.5em}
\end{table}

\section{Problem Definition}
\label{sec:problem_definition}

\paragraph{Definition.} Conceptualizing an event-centric CSKB to derive abstract commonsense knowledge comprises two steps~\cite{DBLP:journals/corr/abs-2206-01532}: event conceptualization and triple conceptualization.

Denote the triples in the original CSKB as $D_o=\{(h_o,r,t)|h_o \in H_o, r \in R, t \in T\}$, where $H_o$, $R$, and $T$ are the set of heads, relations, and tails in the original CSKB. 
The first step only operates on head events without considering the context in $r$ and $t$. 
The goal of event conceptualization is to produce conceptualized head event $h_a$ from the original head $h_o$ to represent an abstraction of $h_o$.
In the second step, the task is to verify whether the conceptualized head $h_a$ still makes sense in the context of $r$ and $t$, as $r$ and $t$ will further restrict the level of abstractness in $h_a$.
As shown in Figure~\ref{fig:Introduction_demo}, conceptualizing \textit{watch football game} to \textit{observe} is wrong within the context of having \textit{feel relaxed} as a result.
Plausible $(h_a,r,t)$ triples will be considered as valid abstract commonsense knowledge.

Specifically, in the first step, there are two ways of conceptualizing head events alone: a \textit{retrieval-based discriminative} way and a \textit{generative} way. 
The retrieval-based discriminative paradigm identifies and links a component $i$ in $h_o$ to a concept $c$ in a concept taxonomy $C$ to form a conceptualization $h_a$ by replacing $i$ with $c$.
The model needs to verify whether $h_a$ is a valid conceptualization of $h_o$.
The generative paradigm aims to generate a $h_a$ directly given $h_o$ and the designated component $i$ in $h_o$.

Formally, denote the annotated dataset in the first step, event conceptualization, as $D^l_h=\{ (h_o, h_a, y) | h_o \in H_o, h_a \in H_a, y \in \{0,1\}\}$, 
where $h_o$ is an original head event without conceptualization, $h_a$ is a corresponding conceptualization of $h_o$, and $y$ is the human-annotated label indicating whether such a conceptualization is plausible or not.
The labeled dataset in the second step, triple conceptualization, is denoted as $D^l_t=\{ (h, r, t,y) | h \in H_a, r \in R,t \in T, y \in \{0,1\}\}$, 
where $h$ is a conceptualized head event from the first step, $r$ and $t$ are a relation and a tail from the original CSKB accompanied with the corresponding original head $h_o$, and $y$ is the human-annotated label indicating whether such abstract commonsense knowledge, in the form of a conceptualized triple, is plausible or not. 
Besides labeled datasets, unlabeled datasets are defined similarly as $D_h^u$ and $D_t^u$ only with the difference that labels $y$ are missing.
Thus, the task objective for discriminative event conceptualization is to determine whether a $h_o$ can be properly abstracted using $h_a$, where $h_a$ is derived by replacing a component $i \subset h_o$ with its linked concept $c$ from a concept taxonomy $C$. 
The task objective for generative event conceptualization is to generate $h_a$ directly from $h_o$ with text generation models.
For the triple conceptualization task, the objective is to distinguish whether a conceptualized triple $(h_a, r, t)$, representing abstract commonsense knowledge, is plausible or not.

\paragraph{Dataset. }
To study conceptualization over CSKBs, we use the AbstractATOMIC dataset provided by~\citet{DBLP:journals/corr/abs-2206-01532} as the benchmark.
In AbstractATOMIC, ATOMIC is used as the original CSKB.
And the event conceptualization adopts a \textit{discriminative} way, where a syntactic parsing schema is defined to identify the components $i$ in $h_o$ to be heuristically linked to concept taxonomies Probase~\cite{DBLP:conf/sigmod/WuLWZ12} and WordNet~\cite{DBLP:journals/cacm/Miller95} to form conceptualized $h_a$.
Such a heuristic can produce over 32 times more candidate conceptualized head events and over 10 times more conceptualized triples compared with the original ATOMIC, as the number of retrieved concepts from the concept taxonomy $C$ can be manually controlled to acquire a large number of conceptualizations. 
Triple conceptualization is defined as predicting the plausibility of the triples whose head is conceptualized.
Only 131K (26\%) conceptualizations of 7K (45\%) ATOMIC head events and 81K (1.3\%) conceptualized triples are manually annotated as $D_h^l$ and $D_t^l$, while others remain unlabeled $D_h^u$ and $D_t^u$.
The \textit{trn/dev/tst} partition follows the same split as in the original ATOMIC. 
Statistics and more detailed explanations of AbstractATOMIC are shown in Table~\ref{tab:AbstractAtomic_statistics} and Appendix~\ref{sec:appendix_dataset_statistics}.

\section{CAT Framework}
\label{sec:CAT_framework}

This section introduces our proposed Contextualized ConceptualizAtion and InsTantiation (CAT) framework for conceptualizing commonsense knowledge bases and acquiring abstract commonsense knowledge.
An overview is presented in Figure~\ref{fig:CAT_overview}.
Our motivation is two-fold:
first, adding instantiation after conceptualization to form a cycle can strongly benefit two conceptualization tasks simultaneously.
On the one hand, instantiating conceptualized triple relies on the correctness of event conceptualization.
On the other hand, properly conceptualized triples can benefit event conceptualization via instantiation by providing more context brought by $(r,t)$. 
Second, to address the lack of annotations, we resort to pseudo labeling, a typical semi-supervised learning approach to automatically assign pseudo labels to the vast majority of unlabeled data using a teacher model.

Following~\citet{DBLP:journals/corr/abs-2206-01532}, we study the retrieval-based discriminative paradigm of event conceptualization and leave the generative paradigm as an intrinsic evaluation. 
In CAT, we unify event conceptualization and triple conceptualization into one cycle and make them mutually benefit each other through instantiation and conceptualization. 
Our framework can be summarized into four steps:

\noindent (1) Train teacher models for both event conceptualization and triple conceptualization on the labeled dataset $D_h^l$ and $D_t^l$, respectively. Use the two teachers to assign pseudo labels to unlabeled datasets.

\noindent (2) Conduct alternative conceptualization or instantiation on labeled and pseudo-labeled data.

\noindent (3) Bootstrap (aggregate) the alternative concepts and instances in the second step using natural language prompt templates and train student models on both labeled and pseudo-labeled data.

\noindent (4) Use the student models to refine the pseudo labels and then re-train the student models.

\subsection{Teacher Model Training}
\label{sec:teacher_model_training}

Two teacher models on both event and triple conceptualization tasks are trained separately on the labeled dataset $D_h^l$ and $D_t^l$.
As both tasks are inherently text/triple classification, we adopt
 KG-BERT~\cite{DBLP:journals/corr/abs-1909-03193} as the skeleton of our models.
The event conceptualization model determines whether $h_a$ is a valid conceptualization of $h_o$, and the triple conceptualization model determines whether a conceptualized triple $(h_a,r,t)$ is plausible or not.
The two models $\theta$ are trained on annotated examples $x_i$ with a cross-entropy loss (Eq.~\ref{eq:BCE_loss}) and used to provide pseudo labels to instances from the unlabeled datasets $D_h^u$ and $D_t^u$.
Two thresholds, $T^+$ and $T^-$, are set to determine the pseudo labels of unlabeled examples with high confidence. 
Examples with a pseudo-labeled score higher than $T^+$ will be labeled $y_i=1$, and those lower than $T^-$ will be labeled $y_i=0$. 
The rest will be discarded. 
\vspace{-1em}

\begin{equation}
    \label{eq:BCE_loss}
    L(x_i,\theta)=-\sum_{i=1}^{|x|}y_i\log(\theta(x_i)) 
\end{equation}

\subsection{Alternative Conceptualization and Instantiation}

According to~\citet{murphy2004big}, when humans learn a new concept, we pre-extract similar known concepts in our minds and infer possibly equivalent unknown concepts on the fly.
Inspired by this theory, we retrieve additional abstract concepts or instantiated events to help discriminate conceptualizations and abstract commonsense knowledge.
For event conceptualization, 
we retrieve some alternative possible conceptualizations of $h_o$ to accompany the learning of $h_a$.
Additional conceptualizations of $h_o$ from both labeled and pseudo-labeled examples are predicted again by the teacher model and ranked according to their plausibility score prediction.
And top $m$ conceptualizations are retrieved with $m$ being a hyperparameter to control the number of retrievals.
For triple conceptualization, 
we perform instantiation in cascade to instantiate $c$ to some concrete instances to assist the learning process.
Possible instantiations of $c$ are extracted from annotated and pseudo-labeled event conceptualizations by searching for conceptualized events $h'_a \in H_a$ other than $h_a$ with $c$ as the concept and extracting their corresponding instances $i \subset h'_a$.
Similarly, the instances are then scored by the teacher model, and the top $n$ of them are retrieved.
Intuitively, alternative event conceptualizations can serve as hints for discriminating the correctness of the target conceptualization, and instantiations can carry additional contextualized information to help verify the plausibility of a conceptualized triple, which meets the objective of deriving abstract commonsense knowledge that is context-sensitive.

\begin{table*}[t]
\small
\centering
\begin{tabular}{l|l|cc|cc}
\toprule
\multirow{2}{*}{Framework} & \multirow{2}{*}{Backbone PTLM / Method} & \multicolumn{2}{c|}{Event Conceptualization} & \multicolumn{2}{c}{Triple Conceptualization} \\
\cmidrule(l){3-6}
 &  & Validation & Testing & Validation & Testing \\ 
 \midrule
\multirow{14}{*}{\begin{tabular}[c]{@{}l@{}}Supervised\\ Learning\end{tabular}} & BERT-base \scriptsize{\textit{110M}} & 82.4$_{\pm\text{0.05}}$
 & 82.5$_{\pm\text{0.31}}$ & 71.2$_{\pm\text{0.58}}$ & 72.6$_{\pm\text{0.71}}$ \\
& BERT-large \scriptsize{\textit{340M}} & 82.8$_{\pm\text{0.48}}$ & 83.1$_{\pm\text{0.80}}$ & 72.4$_{\pm\text{0.01}}$ & 73.7$_{\pm\text{0.00}}$ \\
& BART-base \scriptsize{\textit{139M}} & 83.8$_{\pm\text{0.28}}$ & 84.4$_{\pm\text{0.32}}$ & 72.0$_{\pm\text{0.09}}$ & 72.6$_{\pm\text{0.15}}$ \\
& BART-large \scriptsize{\textit{406M}} & 85.0$_{\pm\text{0.13}}$ & 85.2$_{\pm\text{0.22}}$ & 74.5$_{\pm\text{0.13}}$ & 76.2$_{\pm\text{0.19}}$ \\
& RoBERTa-base \scriptsize{\textit{110M}} & 84.1$_{\pm\text{0.04}}$ & 84.5$_{\pm\text{0.19}}$ & 72.2$_{\pm\text{0.00}}$ & 74.1$_{\pm\text{0.00}}$ \\
& RoBERTa-large \scriptsize{\textit{340M}} & 85.2$_{\pm\text{0.24}}$ & 85.5$_{\pm\text{0.02}}$ & 75.3$_{\pm\text{0.00}}$ & 76.9$_{\pm\text{0.01}}$ \\
& DeBERTa-v3-base \scriptsize{\textit{214M}} & 85.1$_{\pm\text{0.08}}$ & 85.8$_{\pm\text{0.07}}$ & 73.9$_{\pm\text{0.10}}$ & 75.9$_{\pm\text{0.04}}$ \\
& DeBERTa-v3-large \scriptsize{\textit{435M}} & \underline{85.8$_{\pm\text{0.05}}$} & \underline{86.2$_{\pm\text{0.15}}$} & \underline{76.9$_{\pm\text{0.03}}$} & \underline{78.0$_{\pm\text{0.02}}$} \\
& ELECTRA-base \scriptsize{\textit{110M}} & 85.4$_{\pm\text{0.05}}$ & 85.8$_{\pm\text{0.02}}$ & 74.3$_{\pm\text{0.27}}$ & 76.2$_{\pm\text{0.12}}$ \\
& ELECTRA-large \scriptsize{\textit{340M}} & 84.7$_{\pm\text{0.47}}$ & 85.3$_{\pm\text{0.38}}$ & 75.6$_{\pm\text{0.01}}$ & 77.9$_{\pm\text{0.06}}$ \\
\cmidrule{2-6}
& GPT2-base \scriptsize{\textit{117M}} & 60.0$_{\pm\text{0.06}}$ & 59.1$_{\pm\text{0.14}}$ & 52.8$_{\pm\text{0.14}}$ & 55.9$_{\pm\text{0.11}}$ \\
& GPT2-medium \scriptsize{\textit{345M}} & 61.2$_{\pm\text{0.11}}$ & 60.3$_{\pm\text{0.08}}$ & 54.6$_{\pm\text{0.17}}$ & 57.4$_{\pm\text{0.09}}$ \\
& GPT2-large \scriptsize{\textit{774M}} & 64.1$_{\pm\text{0.05}}$ & 62.7$_{\pm\text{0.08}}$ & 60.5$_{\pm\text{0.11}}$ & 59.8$_{\pm\text{0.06}}$ \\
& GPT2-XL \scriptsize{\textit{1558M}} & 64.2$_{\pm\text{0.19}}$ & 63.6$_{\pm\text{0.22}}$ & 62.2$_{\pm\text{0.08}}$ & 61.5$_{\pm\text{0.10}}$ \\
\midrule
\multirow{5}{*}{\begin{tabular}[c]{@{}l@{}}Semi-Supervised\\ Learning\end{tabular}} & UDA (TF-IDF) &83.6$_{\pm\text{0.29}}$ &83.6$_{\pm\text{0.24}}$ & 75.8$_{\pm\text{1.26}}$ & 76.8$_{\pm\text{1.34}}$ \\ 
& UDA (back-trans.) &83.4$_{\pm\text{0.27}}$ & 83.6$_{\pm\text{0.24}}$ & 75.8$_{\pm\text{1.25}}$ & 76.8$_{\pm\text{1.34}}$ \\
& Noisy-Student & 86.4$_{\pm\text{0.05}}$ & 86.5$_{\pm\text{0.09}}$ & 75.4$_{\pm\text{0.64}}$ & 76.7$_{\pm\text{0.59}}$ \\
& PseudoReasoner (BERT-base) & 83.3$_{\pm\text{0.11}}$ & 84.0$_{\pm\text{0.24}}$ & 73.0$_{\pm\text{0.14}}$ & 74.1$_{\pm\text{0.33}}$ \\
& PseudoReasoner (RoBERTa-large) & \underline{86.6$_{\pm\text{0.25}}$} & \underline{86.7$_{\pm\text{0.33}}$} & \underline{76.3$_{\pm\text{0.12}}$} & \underline{77.2$_{\pm\text{0.21}}$} \\
\midrule
\multirow{10}{*}{\begin{tabular}[c]{@{}l@{}}\textbf{CAT}\\ \textbf{\textit{(Semi-Supervised)}}\end{tabular}} & BERT-base \scriptsize{\textit{110M}} & 87.1$_{\pm\text{0.06}}$ & 87.4$_{\pm\text{0.11}}$ & 74.3$_{\pm\text{0.26}}$ & 76.3$_{\pm\text{0.38}}$ \\
& BERT-large \scriptsize{\textit{340M}} & 87.7$_{\pm\text{0.16}}$ & 88.0$_{\pm\text{0.19}}$ & 75.8$_{\pm\text{0.23}}$ & 77.8$_{\pm\text{0.36}}$ \\
& BART-base \scriptsize{\textit{139M}} & 88.2$_{\pm\text{0.09}}$ & 88.2$_{\pm\text{0.09}}$ & 75.7$_{\pm\text{0.09}}$ & 78.0$_{\pm\text{0.14}}$ \\
& BART-large \scriptsize{\textit{406M}} & 88.6$_{\pm\text{0.07}}$ & 88.7$_{\pm\text{0.10}}$& 77.2$_{\pm\text{0.12}}$ & 79.0$_{\pm\text{0.14}}$ \\
& RoBERTa-base \scriptsize{\textit{110M}} & 88.4$_{\pm\text{0.12}}$ & 88.3$_{\pm\text{0.08}}$ & 76.9$_{\pm\text{0.16}}$ & 78.0$_{\pm\text{0.19}}$ \\
& RoBERTa-large \scriptsize{\textit{340M}} & 89.0$_{\pm\text{0.15}}$ & 88.8$_{\pm\text{0.20}}$ & 78.2$_{\pm\text{0.08}}$ & 79.4$_{\pm\text{0.14}}$ \\
& DeBERTa-v3-base \scriptsize{\textit{214M}} & 88.8$_{\pm\text{0.12}}$ & 88.9$_{\pm\text{0.08}}$ & 77.5$_{\pm\text{0.10}}$ & 79.9$_{\pm\text{0.07}}$ \\
& DeBERTa-v3-large \scriptsize{\textit{435M}} & \textbf{\underline{89.1$_{\pm\text{0.05}}$}} & \textbf{\underline{89.2$_{\pm\text{0.14}}$}} & \textbf{\underline{78.7$_{\pm\text{0.16}}$}} & \textbf{\underline{80.0$_{\pm\text{0.33}}$}} \\
& ELECTRA-base \scriptsize{\textit{110M}} & 88.7$_{\pm\text{0.10}}$ & 88.9$_{\pm\text{0.10}}$ & 74.9$_{\pm\text{0.15}}$ & 75.5$_{\pm\text{0.40}}$ \\
& ELECTRA-large \scriptsize{\textit{340M}} & 88.6$_{\pm\text{0.77}}$ & 88.5$_{\pm\text{0.70}}$ & 74.9$_{\pm\text{0.15}}$ & 75.5$_{\pm\text{0.40}}$ \\
\bottomrule
\end{tabular}
\caption{Performance (\%) by our CAT framework on the discriminative event conceptualization and triple conceptualization tasks. 
We report the average AUC score and standard deviation across experiments with three random seeds.
The best performances within each framework are underlined, and the best among all models are bold-faced. 
}
\label{tab:CAT_performance}
\end{table*}

\subsection{Prompt Aggregation}
\label{sec:prompt_aggregaation}
We then bootstrap the retrieved alternative conceptualizations/instantiations via natural language prompts.
Here bootstrap~\cite{carey2004bootstrapping} can be understood as binding the alternative retrievals and the target concept/triple together to strengthen the discrimination of the target concept/triple.
As shown in Figure~\ref{fig:CAT_overview} step (3), the initially given input and retrieved concepts/instances are concatenated via human-defined prompts for both conceptualization tasks.
Alternative concepts/instances are sorted in the order of their plausibility score ranking.
Two student models $\mathcal{S}_h$ and $\mathcal{S}_t$ for both tasks are trained using the modified text with such prompts as inputs.
They are expected to learn the bootstrapping connectionism between the target and the additional retrievals we provided.
More detail about the prompt design is in Appendix~\ref{sec:appendix_prompt_design}.

\begin{table*}[t]
\small
\centering
\begin{tabular}{@{}l|cc|cc|cc|cc|cc|cc@{}}
\toprule
\multirow{2}{*}{Training Data} & \multicolumn{2}{c|}{BLEU-1} & \multicolumn{2}{c|}{BLEU-2} & \multicolumn{2}{c|}{METEOR} & \multicolumn{2}{c|}{ROUGE-L} & \multicolumn{2}{c|}{CIDEr} & \multicolumn{2}{c}{\textbf{Human}} \\ 
\cmidrule(l){2-13} 
& Dev & Test & Dev & Test & Dev & Test & Dev & Test & Dev & Test & Dev & Test \\ 
\midrule
$D^l_h + D^u_{0.95}$ & \textbf{73.0} & \underline{71.1} & \textbf{70.2} & 63.0 & \textbf{48.1} & \textbf{47.1} & \textbf{71.4} & \underline{70.7} & \textbf{63.6} & \underline{66.9} & \textbf{92.8} & \textbf{93.3} \\
$D^l_h + D^u_{0.9}$ & \underline{71.3} & \textbf{71.9} & 65.2 & \underline{63.8} & \underline{45.7} & \underline{46.7} & \underline{69.8} & \textbf{71.3} & \underline{63.4} & \textbf{67.9} & \underline{90.5} & \underline{91.0} \\
$D^l_h + D^u_{0.8}$ & 68.2 & 68.4 & \underline{65.9} & \textbf{64.0} & 44.8 & 44.0 & 66.6 & 66.7 & 60.0 & 62.0 & 86.0 & 85.7 \\
$D^l_h + D^u_{0.7}$ & 66.5 & 67.2 & 57.2 & 62.6 & 43.0 & 43.4 & 65.9 & 65.8 & 60.4 & 61.2 & 79.0 & 80.3 \\
$D^l_h + D^u_{0.5}$ & 64.9 & 62.4 & 58.3 & 51.1 & 41.2 & 40.9 & 63.8 & 63.0 & 58.2 & 59.4 & 74.5 & 79.0 \\
\midrule
$D^l_h$ & 67.6 & 65.3 & 56.8 & 53.1 & 43.5 & 43.1 & 65.7 & 66.6 & 60.2 & 60.9 & 70.0 & 81.5 \\
Zero-Shot & 20.2 & 17.0 & 6.80 & 4.11 & 5.80 & 4.70 & 3.80 & 3.00 & 1.90 & 1.60 & 15.0 & 11.5 \\
\bottomrule
\end{tabular}
\caption{Performance (\%) of GPT2 (XL) on the generative event conceptualization task. $D_h^l$ stands for annotated labeled data, and $D^u$ stands for the data acquired by CAT. The underfoot value indicates the threshold for selecting plausible pseudo labels. The best performances are bold-faced, and the second-best ones are underlined.}
\label{tab:head_concept_generation_performance}
\end{table*}

\subsection{Pseudo-Label Refinement}

All pseudo labels, initially derived by a teacher model trained on the original labeled dataset, are re-labeled according to the plausibility score predicted by our newly enhanced student models $\mathcal{S}_h$ and $\mathcal{S}_t$.
Similar to the teacher model, two thresholds, $T^+$ and $T^-$, are applied to distinguish positive and negative examples for both tasks.
In addition, negative labels are assigned to triples whose conceptualized head events are predicted as wrong conceptualizations by $\mathcal{S}_h$, as wrong conceptualizations will not yield plausible abstract commonsense knowledge.

\subsection{Application and Evaluation of CAT}
\label{sec_application_evaluation_CAT}

The resulting models of CAT include an event conceptualization model and a triple conceptualization model, both fine-tuned on the refined pseudo labels and the labeled data. 
These two models can be used to conceptualize ATOMIC to a larger commonsense knowledge base on a more abstract level.
We further conduct intrinsic evaluations on the acquired event conceptualization model under a generative event conceptualization paradigm and extrinsic evaluations on the resulting conceptualized CSKB with commonsense inference modeling task (COMET; \citet{DBLP:conf/acl/BosselutRSMCC19}) in Section~\ref{experiments}.
Here we select COMET as the representative because it is a general commonsense model that can be applied to various downstream commonsense reasoning tasks such as SocialIQA~\cite{DBLP:conf/emnlp/SapRCBC19}, self-talk~\cite{DBLP:conf/emnlp/ShwartzWBBC20}, and CSKB completion~\cite{DBLP:conf/aaai/MalaviyaBBC20}.
\wq{Meanwhile, generative event conceptualization enables performing automatic conceptualization scalably.
Both are important applications and evaluations of CAT.}

\section{Experiments}
\label{experiments}

We conduct conceptualization experiments using CAT in Section~\ref{sec:exp_cskb_concept} and generative experiments as evaluations in Section~\ref{sec:exp_comet}.
These experiments demonstrate that CAT has a strong capability in conceptualizing CSKBs, and better conceptualization modeling can help populate more novel and diverse commonsense knowledge and thus help commonsense modeling (COMET).

\subsection{CSKB Conceptualization}
\label{sec:exp_cskb_concept}
\paragraph{Baselines.}

We collectively introduce the baselines for both event and triple conceptualization tasks, as they are inherently classification tasks.
AUC is used as the evaluation metric.
Under a supervised learning setting, we apply KG-BERT~\cite{DBLP:journals/corr/abs-1909-03193} model with BERT~\cite{DBLP:conf/naacl/DevlinCLT19}, BART~\cite{DBLP:conf/acl/LewisLGGMLSZ20}, RoBERTa~\cite{DBLP:journals/corr/abs-1907-11692}, DeBERTa~\cite{DBLP:conf/iclr/HeLGC21,DBLP:journals/corr/abs-2111-09543}, and ELECTRA~\cite{DBLP:conf/iclr/ClarkLLM20} as the backbone language models.
We also attempt to leverage supervised generative language models as baselines. 
GPT2~\cite{radford2019language} models are trained with a text generation objective only on positive examples, and we use perplexity as the prediction scores to calculate AUC.
For the semi-supervised learning baselines, we leverage UDA~\cite{DBLP:conf/nips/XieDHL020}, NoisyStudent~\cite{DBLP:conf/cvpr/XieLHL20}, and PseudoReasoner~\cite{DBLP:journals/corr/abs-2210-07988} with RoBERTa-large being the backbone model. 
Additional explanations can be found in Appendix~\ref{sec:appendix_baseline_training_detail}.

\paragraph{Discriminative Results.}
The results for both tasks are presented in Table~\ref{tab:CAT_performance}. 
Under a supervised learning setting, 
KG-BERT family mostly performs better on both tasks than GPT2 due to the fact that GPT2 is only fine-tuned on positive examples and thus cannot learn from negative examples that contain wrong conceptualizations and implausible abstract commonsense knowledge.
As for the semi-supervised learning setting, previous SSL baselines are rather limited in improving the performance against supervised learning.
The best PseudoReasoner only improves by 0.5\% and 0.3\% on the test set for both tasks compared with supervised RoBERTa-large models. 
Instead, models trained with CAT can outperform all other training methodologies. 
Comparing the test set performance with PseudoReasoner, small backbone models (BERT-base) can improve by 3.4\% and 2.2\%, and large models (RoBERTa-large) can be improved by 2.1\% and 2.2\%.
This shows pipelining two-step conceptualizations as a loop and leveraging our proposed bootstrapping-based method can yield a larger performance gain compared with simply applying a semi-supervised learning strategy.
Due to limited space, ablation studies on framework components and the semi-supervised learning paradigm of CAT are conducted in Appendix~\ref{sec:ablation_study}.
For example, the results indicate that bootstrapping alternative conceptualization and instantiation plays the most important role in assisting learning conceptualization among all components of CAT.
Additional results and a computational cost study can be found in Appendix~\ref{sec:appendix_verification_result} and Appendix~\ref{sec:appendix_computational_cost}.

\begin{table*}[t]
\small
\centering
\begin{tabular}{@{}l|cc|cc|cc|cc|cc|cc|cc@{}}
\toprule
\multirow{2}{*}{Training Data} & \multicolumn{2}{c|}{BLEU-1} & \multicolumn{2}{c|}{BLEU-2} & \multicolumn{2}{c|}{BLEU-3} & \multicolumn{2}{c|}{BLEU-4} & \multicolumn{2}{c|}{METEOR} & \multicolumn{2}{c|}{ROUGE-L} & \multicolumn{2}{c}{CIDEr} \\ \cmidrule(l){2-15}
 & Dev & Test & Dev & Test & Dev & Test & Dev & Test & Dev & Test & Dev & Test & Dev & Test \\
\midrule
Zero-Shot & 5.42 & 4.89 & 1.84 & 1.51 & 0.65 & 0.52 & 0.26 & 0.21 & 6.50 & 5.70 & 6.40 & 5.90 & 1.60 & 1.20 \\
ATOMIC (subset) & 38.1 & 38.1 & 25.4 & 25.7 & 18.7 & 18.8 & 15.5 & 15.7 & 14.9 & 14.9 & 33.0 & 33.2 & 27.6 & 27.8 \\
\midrule
$+D^l_t$ & 38.1 & 38.5 & 24.8 & 25.5 & 17.8 & 18.4 & 14.7 & 15.2 & 15.3 & 15.6 & 33.1 & 33.7 & 26.8 & 27.3 \\
\hspace{3mm}$+\text{Finetune}$ & 38.6 & 39.0 & 25.8 & 26.6 & 18.9 & 19.7 & 15.7 & 16.4 & 15.1 & 15.4 & 33.6 & 34.4 & 28.8 & 30.0 \\
$+D^{u}_{\text{Abs.ATM.}}$ & 40.0 & 40.3 & 27.1 & 27.8 & 20.0 & 20.8 & 16.5 & 17.5 & 16.1 & 16.3 & 35.3 & 35.7 & 31.6 & 31.7 \\
\hspace{3mm}$+\text{Finetune}$ & 40.1 & 40.5 & 27.1 & 27.8 & 20.1 & 20.8 & 16.7 & 17.4 & 16.2 & 16.4 & 35.4 & 35.9 & 31.8 & 31.7 \\
$+D^l_t+D^{u}_{\text{Abs.ATM.}}$ & 40.2 & 40.6 & 26.2 & 27.4 & 19.0 & 20.4 & 15.1 & 16.8 & 16.3 & 16.5 & 35.0 & 35.4 & 31.0 & 31.3 \\
\hspace{3mm}$+\text{Finetune}$ & 40.0 & 40.4 & 26.0 & 26.9 & 18.7 & 19.7 & 15.0 & 16.1 & 16.3 & 16.4 & 35.0 & 35.4 & 30.3 & 30.7 \\
\midrule
$+D^u_{\text{CAT}}$ & \textbf{41.2} & 41.9 & \textbf{28.1} & \textbf{29.0} & \textbf{20.7} & \textbf{21.5} & \textbf{16.5} & \textbf{17.8} & \textbf{16.6} & 16.9 & 35.9 & 36.5 & \textbf{33.4} & 33.7 \\
\hspace{3mm}$+\text{Finetune}$ & 41.1 & \textbf{42.0} & 28.0 & 29.0 & 20.4 & 21.5 & 16.4 & 17.6 & 16.6 & \textbf{17.0} & \textbf{36.0} & \textbf{36.8} & 33.2 & \textbf{33.8} \\
$+D^l_t+D^u_{\text{CAT}}$ & 39.9 & 40.5 & 26.2 & 27.4 & 19.3 & 20.6 & 16.0 & 17.4 & 16.0 & 16.2 & 35.0 & 35.4 & 30.8 & 31.3 \\
\hspace{3mm}$+\text{Finetune}$ & 40.4 & 41.0 & 26.6 & 27.6 & 19.5 & 20.7 & 16.1 & 17.1 & 16.2 & 16.5 & 35.4 & 35.8 & 31.3 & 31.5 \\
\bottomrule
\end{tabular}
\caption{Performances (\%) of GPT2 (XL) on commonsense inference modeling task (COMET). 
$D^l_t$ stands for annotated abstract triples, and $D^u_{\text{CAT}}$ stands for abstract triples acquired by CAT. 
$D^{u}_{\text{Abs.ATM.}}$ contains triples that are pseudo-labeled by a supervised RoBERTa discriminator, as done by~\citet{DBLP:journals/corr/abs-2206-01532}.
The best performances are bold-faced.
Finetune refers to fine-tuning back on the ATOMIC subset.}
\label{tab:triple_tail_generation_performance}
\vspace{-0.1in}
\end{table*}

\subsection{Application and Evaluation of CAT}
\label{sec:exp_comet}

As CAT is a framework for acquiring conceptualized commonsense knowledge, including both conceptualized head events (from $h_o$ to $h_a$) and abstract commonsense triples $(h_a, r, t)$, 
we assess these pseudo-labeled outcomes via two generative tasks with various threshold tuning as evaluations.
\paragraph{Generative Event Conceptualization.}
\label{exp_generative_conceptualization}

To intrinsically evaluate the effectiveness of CAT's event conceptualization, 
we use the acquired conceptualized head events as training data to learn a generative event conceptualizer.
Specifically, the models are trained with instance-conceptualizations pairs in the format of ``\textit{<instance> is an instance of <concept>}''.
At the evaluation phase, the model is prompted with ``\textit{<instance> is an instance of} \texttt{[GEN]}'' where \textit{<instance>} is the instance to be conceptualized and \texttt{[GEN]} is the generation token. 
We then retrieve the top-1 generation and compare it against the target set from the evaluation dataset to compute four NLG metrics, as listed in Appendix~\ref{sec:appendix_application_CAT_settings}.
These scores can be regarded as an approximation of the top-1 generations' recall.
Additionally, we uniformly sample 500 generations from each evaluation split and conduct expert annotations on the plausibility of each conceptualization to ensure that out-of-domain concepts can be properly evaluated.
The experts are asked to determine whether each top-1 generation is indeed a plausible conceptualization or not, such that the top-1 generations' precision is reflected.
Thus, current evaluation measures jointly evaluate the top-1 generations' precision and recall, which makes it robust and non-easy to be impacted by repetition problems~\cite{DBLP:conf/acl/LiRKWBCW20}.
Zero-shot GPT2 and GPT2 fine-tuned on the originally labeled event conceptualizations in $D_h^l$ are used as baselines.
We also study the effect of the threshold $T^+$ that selects plausible conceptualized heads, where higher thresholds indicate higher plausibility regarded by CAT.
The results are presented in Table~\ref{tab:head_concept_generation_performance}. 
With a relatively high threshold, generators trained on a mixture of pseudo-labeled data by CAT and annotated concepts significantly outperform the baselines in every automated metric. 
A plausible rate of 93.3\% is maximally achieved on the test set, which is 11.8\% higher than the baseline.
Gradually reducing the threshold also decreases the performance, indicating abstract heads with lower plausibility scores can be of poorer quality.
Such results indicate that CAT can produce high-quality event conceptualizations for generative models to learn better conceptualizers without the need to annotate a large number of data.

\paragraph{Commonsense Inference Modeling (COMET).}
The second component of CAT produces triple-level abstract commonsense knowledge.
We evaluate these abstract commonsense triples with a commonsense inference task that generates commonsense tails given heads and relations as inputs, as in COMET~\cite{DBLP:conf/acl/BosselutRSMCC19}.
Following \citet{DBLP:journals/corr/abs-2206-01532}, \wq{we apply the same training and evaluation process to the models.}
The base training data we use are a subset of ATOMIC triples corresponding to those annotated abstract triples in $D^l_t$, which contains 17K (3.7\%) among the original ATOMIC. 
We derive abstract commonsense knowledge using CAT from a subset of $D^u_t$ where the heads correspond to those in the ATOMIC subset to ensure no data leakage, denoted as $D^u_\text{CAT}$.
GPT2 is fine-tuned on the ATOMIC subset, the annotated abstract triples $D^l_t$, the abstract knowledge verified by CAT, or their combinations.
The commonsense generation results are presented in Table~\ref{tab:triple_tail_generation_performance}.
Similar to COMET~\cite{DBLP:conf/acl/BosselutRSMCC19}, all models are evaluated on the original ATOMIC's full validation and testing sets.
The best result is achieved using a mixture of the ATOMIC subset and abstract triples pseudo-labeled by our framework, with 0.95 as the threshold for selecting plausible triples.
This indicates high-quality abstract commonsense triples can indeed provide a more general view of the original commonsense knowledge, thus helping commonsense inference.
Additionally, training with our pseudo-labeled examples outperforms training with those annotated triples in AbstractATOMIC, which also validates the effectiveness of our model that leverages a large amount of unlabeled data.
To further investigate how conceptual knowledge improves commonsense inference modeling, we conduct more empirical analysis in Section~\ref{sec:conceptual_knowledge_COMET}.
Additional experiment results with other thresholds and case studies can be found in Appendix~\ref{sec:appendix_generation_result} and Appendix~\ref{sec:appendix_case_study}, respectively.

\subsection{Number of Retrieved Alternative Conceptualizations and Instantiations.}
\label{sec:number_retrieved_alternatives}
\begin{figure}[t]
    \centering
    \includegraphics[width=0.95\linewidth]{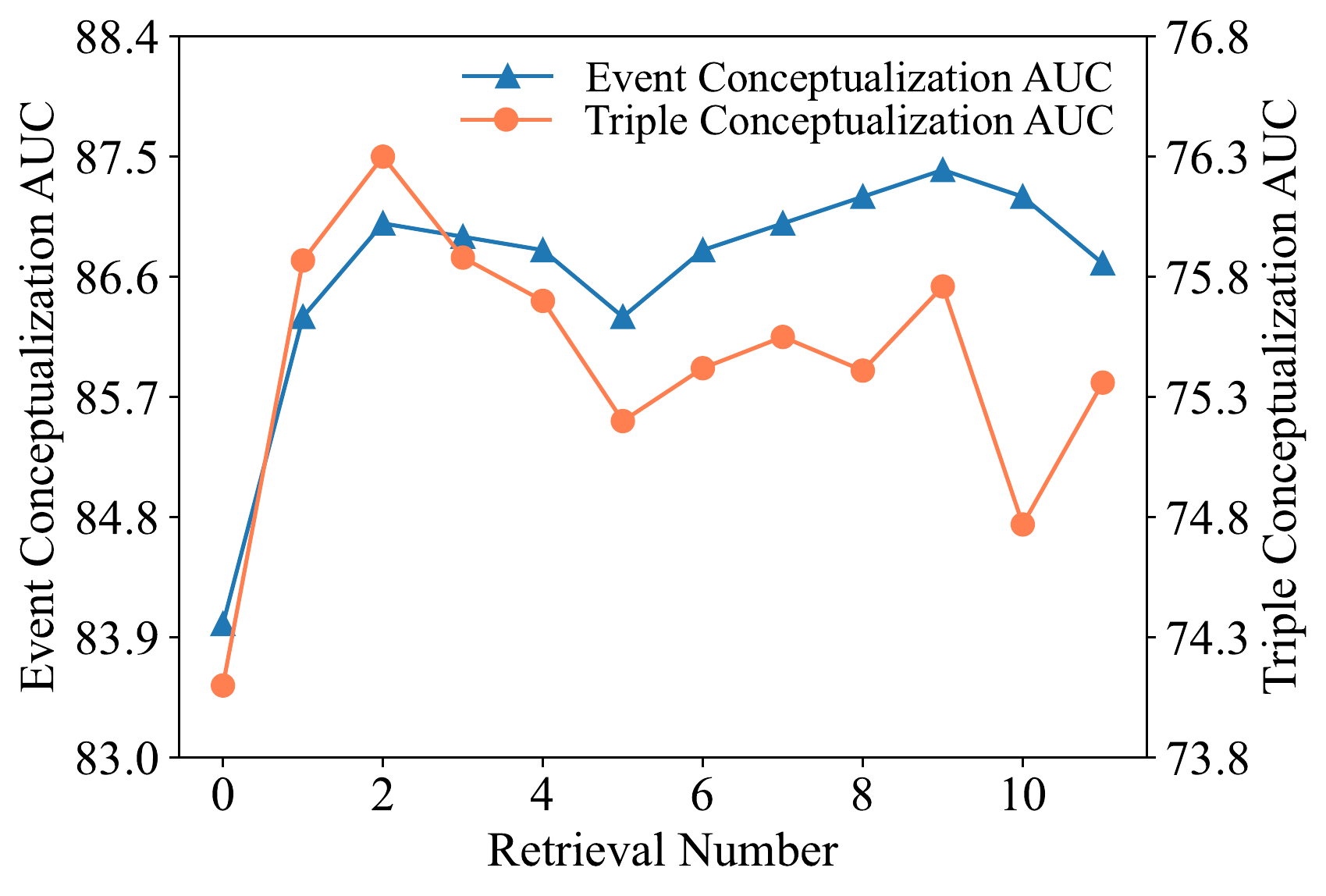}
    \caption{Ablation study on the number of retrieved conceptualizations/instantiations for CAT framework.}
    \label{fig:candidate_number_curve}
    \vspace{-0.1in}
\end{figure}

We then study the ablation of bootstrapping different numbers of alternative conceptualizations/instantiations (denoted as \#retrieval) in our CAT framework. 
For simplicity, when tuning the \#retrieval for one task, the \#retrieval of the other task is fixed at the best value we acquired.
We plot the test AUC score with \#retrieval from 0 to 11 using BERT-base as the backbone model in Figure~\ref{fig:candidate_number_curve}.
\#retrieval=0 refers to training with a simple student-teacher framework without bootstrapping alternative conceptualizations and instantiations.
For event conceptualization, 
the performance generally positively correlates with the number of retrievals, while it starts dropping after 9.
A reversed trend is observed for triple conceptualization, where using only two instances achieves the best performance.
One possible reason is that in triple conceptualization, the retrieved instances are events and much longer than the retrieved concepts in event conceptualization,
and aggregating various alternative events for a triple will cause language models to be less sensitive to the semantics of the original triple~\cite{DBLP:conf/iclr/HoltzmanBDFC20}.

\subsection{The Effect of Abstract Knowledge}
\label{sec:conceptual_knowledge_COMET}

\begin{figure}[t]
    \centering
    \includegraphics[width=0.90\linewidth]{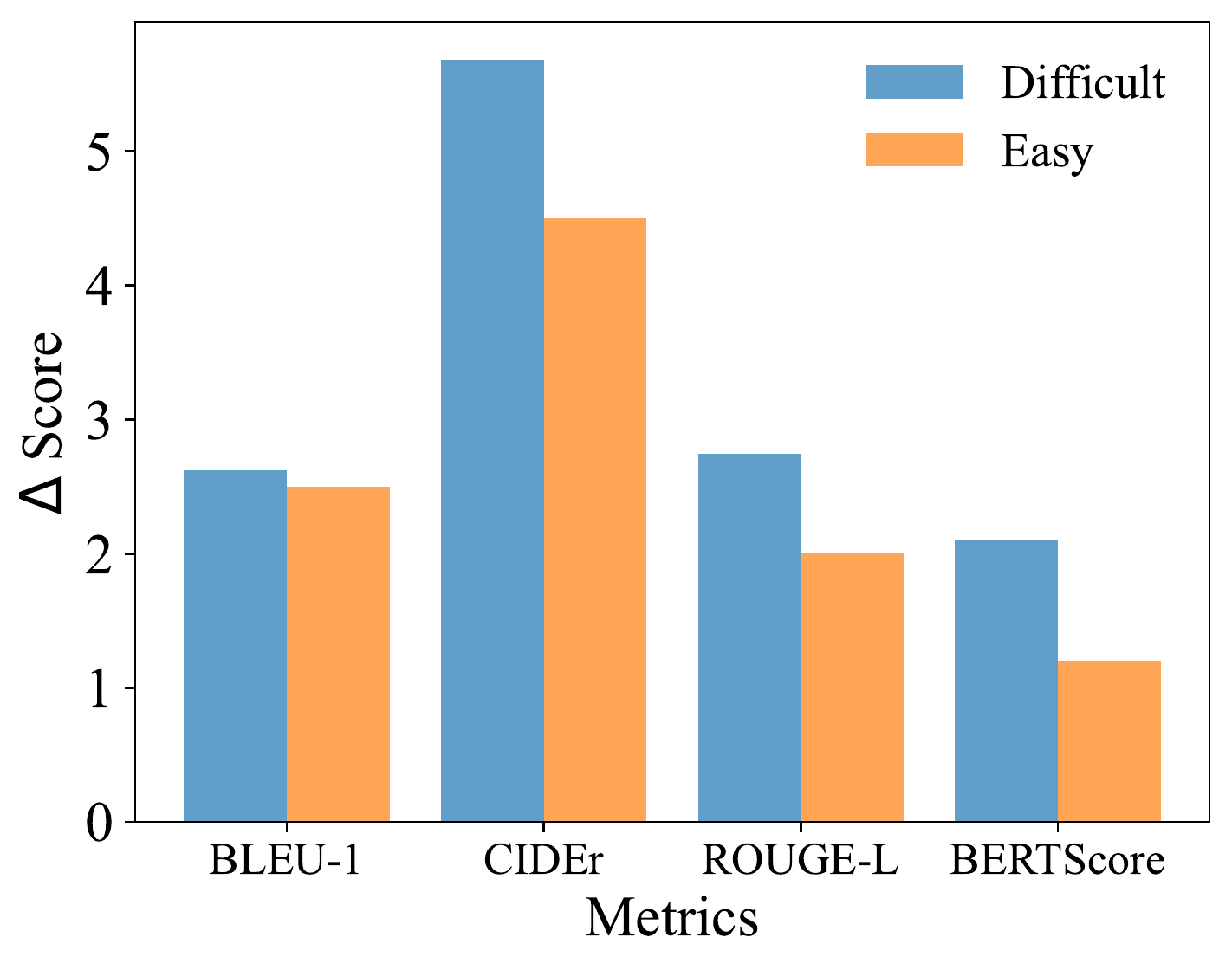}
    \vspace{-1em}
    \caption{Comparison of performance improvement by GPT2 generator trained on the conceptualization-aided ATOMIC subset for two groups of testing head events.}
    \label{fig:generation_similarity_curve}
    \vspace{-0.1in}
\end{figure}

We finally study the effect of abstract commonsense knowledge acquired by CAT by studying the semantic overlaps between training and testing data.
We sort the test set by the BERTScore~\cite{DBLP:conf/iclr/ZhangKWWA20} between each individual testing entry against the whole training set in the original ATOMIC and split them in half to acquire two test groups.
The testing entries with lower BERTScore on the training set indicate a larger semantic shift from the training set~\cite{DBLP:conf/conll/DeutschR21}, 
which is also harder for models to discriminate~\cite{DBLP:conf/cvpr/HsuSJK20}. 
We denote the testing group with a lower BERTScore as ``Difficult'' and the other half as ``Easy''.
The performance gain on the two test set splits between the best conceptualization-aided COMET and the COMET trained on the ATOMIC subset only is reported in Figure~\ref{fig:generation_similarity_curve}.
We can observe that training COMET with abstract commonsense knowledge leads to a larger improvement for harder test examples dissimilar from the original training set, 
indicating that introducing extra abstract commonsense knowledge can help COMET become more generalizable to harder test sets.

\section{Conclusion}
In conclusion, this paper proposes CAT, a semi-supervised learning framework for commonsense reasoning, by leveraging the power of abstract commonsense knowledge.
By achieving state-of-the-art performances in CSKB conceptualization tasks, we remarkably improve modeling commonsense inference, as an important cornerstone of many commonsense reasoning tasks.
Our analysis also demonstrates that high-quality abstract commonsense knowledge can benefit commonsense inference modeling by providing more generalizability on hard commonsense knowledge.
We hope this work can draw insights toward commonsense reasoning from a conceptualization perspective.

\section*{Limitations}

Our framework manually sets thresholds $T^+$ and $T^-$ in pseudo labeling by observations of data quality and hyperparameter searching.
Dynamic threshold tuning~\cite{DBLP:conf/icml/XuSYQLSLJ21} or meta pseudo labels~\cite{DBLP:conf/cvpr/PhamDXL21,DBLP:conf/emnlp/LiZC0SY21} can be implemented to better filter pseudo-labeled examples.
And the thresholds for different tasks can be tuned separately to improve the models' generalizability.

Recently, large generative language models such as GPT3.5~\cite{DBLP:conf/nips/BrownMRSKDNSSAA20} and ChatGPT\footnote{\url{https://chat.openai.com/}}~\cite{DBLP:conf/nips/Ouyang0JAWMZASR22,DBLP:journals/corr/abs-2210-10760} have demonstrated their strong potential on various NLP tasks including probing abstract commonsense knowledge with in-context learning~\cite{DBLP:conf/nips/BrownMRSKDNSSAA20,DBLP:conf/iclr/XieRL022}. 
Due to our limited access, we did not conduct fully-scaled experiments in our paper. 
A short discussion with case studies is provided in Appendix~\ref{sec:appendix_ChatGPT}.

While our framework only operates on AbstractATOMIC as the conceptualization of ATOMIC, it's also worthy of verifying our framework on other CSKBs such as ATOMIC2020~\cite{DBLP:conf/aaai/HwangBBDSBC21}, GLUCOSE~\cite{DBLP:conf/emnlp/MostafazadehKMB20}, ATOMIC10X~\cite{DBLP:conf/naacl/WestBHHJBLWC22}, FolkScope~\cite{DBLP:journals/corr/abs-2211-08316} and eventuality CSKB such as ASER~\cite{DBLP:conf/www/ZhangLPSL20,DBLP:journals/ai/ZhangLPKOFS22} and constructing large conceptualized CSKB benchmarks.
In addition, we only evaluated the power of the acquired abstract commonsense knowledge on the commonsense knowledge generation task (COMET), while other commonsense reasoning tasks remain future works, such as CommonsenseQA~\cite{DBLP:conf/naacl/TalmorHLB19,DBLP:conf/nips/TalmorYBBGCB21}, SocialIQA~\cite{DBLP:conf/emnlp/SapRCBC19}, Winograd Schema Challenge~\cite{DBLP:conf/kr/LevesqueDM12}, PIQA~\cite{DBLP:conf/aaai/BiskZLGC20}, Abductive Commonsense Reasoning~\cite{DBLP:conf/iclr/BhagavatulaBMSH20}, and Winogrande~\cite{DBLP:conf/aaai/SakaguchiBBC20}.

\section*{Ethics Statement}
This paper introduces CAT, a framework for commonsense reasoning via conceptualizing CSKB to acquire abstract commonsense knowledge.
The experiments are conducted on publicly available and well-established datasets that are shared via open-access licenses.
The usage of these datasets in our paper is only for research purposes and is consistent with the datasets' intended usage.
The primary dataset, AbstractATOMIC, largely shares the content with another CSKB, ATOMIC, which is anonymized and desensitized~\cite{DBLP:conf/aaai/SapBABLRRSC19}.
Thus, no data privacy issue is involved.

The potential risks of CAT are relatively low.
Since CAT is trained on AbstractATOMIC, a conceptualization benchmark based on a popular CSKB, ATOMIC, and two concept taxonomies, Proabse and WordNet, it is expected that CAT does not contain any private, offensive, biased, and sensitive information or social, political issues.
The studied tasks all focus on conceptualization or CSKB, which is not likely to generate harmful content, as shown in the case studies in Appendix~\ref{sec:appendix_case_study}.
Thus, we believe that CAT does not yield additional risks. 

\section*{Acknowledgements}
The authors would like to thank the anonymous reviewers for their constructive comments.
The authors of this paper are supported by the NSFC Fund (U20B2053) from the NSFC of China, the RIF (R6020-19 and R6021-20), and the GRF (16211520 and 16205322) from RGC of Hong Kong, the MHKJFS (MHP/001/19) from ITC of Hong Kong and the National Key R\&D Program of China (2019YFE0198200) with special thanks to HKMAAC and CUSBLT. 
We also thank the UGC Research Matching Grants (RMGS20EG01-D, RMGS20CR11, RMGS20CR12, RMGS20EG19, RMGS20EG21, RMGS23CR05, RMGS23EG08).

\bibliographystyle{acl_natbib}
\bibliography{anthology,custom}

\newpage
\appendix

\begin{center}
    {\Large\textbf{Appendices}}
\end{center}

\section{Dataset Description}
\label{sec:appendix_dataset_statistics}

In this section, we introduce more about AbstractATOMIC~\cite{DBLP:journals/corr/abs-2206-01532}, as the primary dataset we experimented with.
AbstractATOMIC is a conceptualized commonsense knowledge benchmark that is built upon ATOMIC~\cite{DBLP:conf/aaai/SapBABLRRSC19}, a popular CSKB in the format of $(h,r,t)$ triples. 
The dataset is entirely in English.
It contains two parts of data: (1) event conceptualization data and (2) abstract knowledge triples conceptualization data.

The event conceptualization data contain conceptualizations for head event instances, where the events are filtered from the original ATOMIC head events. 
Unlike the traditional entity concept taxonomies, where instances are nouns or verb phrases, AbstractATOMIC includes instance candidates that can be either the entire head event or a certain component of an event.
Detailed examples can be found in Appendix~\ref{sec:appendix_case_study}.

The instances within each head event are identified through syntactic parsing by using a parser from the spaCy~\footnote{\href{https://spacy.io/}{https://spacy.io/}} library and matching with five human-defined rules.
After identification, the candidate instances will be heuristically matched against Probase~\cite{DBLP:conf/sigmod/WuLWZ12} and WordNet~\cite{DBLP:journals/cacm/Miller95} via GlossBERT~\cite{DBLP:conf/emnlp/HuangSQH19} to acquire their candidate concepts.
A neural generator based on GPT2, similar to the baseline in this paper, is also trained to generate concepts.
A supervised conceptualization verifier, based on RoBERTa~\cite{DBLP:journals/corr/abs-1907-11692}, is trained as the final gatekeeper to verify the acquired concepts roughly.

\begin{table}[h]
\centering
\small
\begin{tabular}{@{}l|ccc@{}}
\toprule
 & $D^l_h$ & $D^u_h$ & Total \\ 
\midrule
\#Unq. event & 7,196 & 15,165 & 15,388 \\
\#Unq. instance & 7,935 & 20,843 & 21,493 \\
\#Unq. concept & 20,036 & 20,367 & 31,227 \\
\midrule
Avg. \#concept/event & 18.21 & 24.57 & 32.73 \\
Avg. \#concept/instance & 16.51 & 17.88 & 23.43 \\ 
\bottomrule
\end{tabular}
\caption{Additional statistics of the event conceptualization data in AbstractATOMIC (AbsATM). $D^l$ stands for annotated event conceptualizations and $D^u$ are unverified conceptualizations. \# denotes ``number of'', Unq stands for unique, and Avg is average.}
\label{tab:additional_AbsATMconcept_statistics}
\end{table}

Human annotations on the Amazon Mechanical Turk platform are further conducted to acquire annotations on the correctness of 131K conceptualizations of 7K ATOMIC events.
All conceptualizations that are not annotated are regarded as unlabeled data in this paper.
More detailed statistics for the head event conceptualization data can be found in Table~\ref{tab:additional_AbsATMconcept_statistics}.

After acquiring the event conceptualizations by only focusing on head events, abstract commonsense knowledge, in the form of $(h,r,t)$ triple, is collected by connecting conceptualized head event with its non-abstract counterparts (commonsense relations and inference tails) from ATOMIC.
Only the head events contain abstract concepts.
Thus, these abstract triples are more generalized if-then commonsense knowledge that is potentially useful for commonsense reasoning through instantiation.

Human annotations on Amazon Mechanical Turk further verify 81K uniformly sampled abstract triples.
These triples only correspond to 689 unique ATOMIC head events, which makes annotations relatively scarce compared with the scale of unlabeled data.
A supervised RoBERTa-large verifier is trained on the annotated triples to roughly verify abstract triples that are not annotated.
Triples with scores higher than 0.9 are pseudo-labeled as positive ones~\cite{DBLP:journals/corr/abs-2206-01532}.
However, this paper only leverages these pseudo-labeled examples in the commonsense inference generation task (COMET) as baselines.
Only annotated triples are considered hard-labeled for all other tasks concerned.
And triples that are not annotated are treated as unlabeled by default.
The detailed relational distribution of abstract triples is presented in Table~\ref{tab:additional_AbsATMtriple_statistics}.
Examples can be found in Appendix~\ref{sec:appendix_case_study}.

\begin{table}[h]
\centering
\small
\begin{tabular}{@{}l|c|cc|c@{}}
\toprule
Relation & ATOMIC & $D^l_t$ & $D^u_t$ & $D^u_{\text{Abs.ATM.}}$ \\ 
\midrule
xEffect & 78,832 & 12,168 & 938,330 & 451,564 \\
oEffect & 28,351 & 3,526 & 333,845 & 160,207 \\
xWant & 101,249 & 15,312 & 1,170,835 & 543,964 \\
oWant & 43,079 & 5,408 & 484,570 & 227,493 \\
xReact & 62,969 & 8,923 & 510,476 & 288,019 \\
oReact & 26,570 & 3,030 & 224,706 & 126,386 \\
xNeed & 74,272 & 11,733 & 900,429 & 425,060 \\
xAttr & 110,791 & 14,249 & 838,191 & 465,511 \\
xIntent & 45,490 & 6,848 & 519,813 & 259,694 \\
\midrule
Total & 572,053 & 81,197 & 5,921,195 & 2,947,898\\
\bottomrule
\end{tabular}
\caption{Abstract commonsense triple distribution by relations. $D^l$ stands for annotated triples and $D^u$ are unverified triples. $D^u_{\text{Abs.ATM.}}$ stands for abstract triples verified by a supervised RoBERTa-large discriminator, as done by~\citet{DBLP:journals/corr/abs-2206-01532}.}
\label{tab:additional_AbsATMtriple_statistics}
\end{table}

\section{Prompt Design}
\label{sec:appendix_prompt_design}

In this section, we introduce the textual prompts used for training various models.

For event conceptualization, denotes the original event as $h_o$, instance as $i$, target concept to be verified as $c$, and retrieved alternative conceptualizations as $c_{r,1}, c_{r,2}, c_{r,3}, ..., c_{r,m}$.
The prompt for training the teacher model is ``\texttt{[CLS]} $h_o$ \texttt{[SEP]} $c$'', while the one for training the student model is ``\texttt{[CLS]} $h_o$ \texttt{[SEP]} $c$ \texttt{[SEP]} $c_{r,1}$, $c_{r,2}$, $c_{r,3}$, $...$, $c_{r,m}$''.
For the example in Figure~\ref{fig:CAT_overview}, the filled prompt is ``PersonX is on vacation \texttt{[SEP]} relaxing event \texttt{[SEP]} traveling, break, holiday.''
Specifically, special tokens \texttt{<c>} and \texttt{</c>} are used to enclose $i \subset h_o$ within the original event to highlight the instance to be conceptualized.
GPT2 generators use similar prompts, with the difference that \texttt{[SOS]} and \texttt{[EOS]} special tokens are inserted to denote the start and end of the sentence, respectively.

For triple conceptualization, denotes the head, relation, and tail of an abstract commonsense triple as $(h,r,t)$, the abstract concept in the conceptualized head as $c \subset h$, and retrieved instantiations as $e_{r,1}, e_{r,2}, e_{r,3}, ..., e_{r,n}$. 
The prompt for training generally follows the one used by~\citet{DBLP:journals/corr/abs-2206-01532}.
For the teacher model, ``\texttt{[CLS]}, $h_1$, $...$, $h_{|h|}$, \texttt{[SEP]}, $[r]$, \texttt{[SEP]}, $t_1$, $...$, $t_{|t|}$'' is used as the prompt.
Similarly, student models are trained with a prompt ``\texttt{[CLS]}, $h_1$, $...$, $h_{|h|}$ \texttt{[SEP]} $[r]$ \texttt{[SEP]} $t_1$, $...$, $t_{|t|}$ \texttt{[SEP]} $e_{r,1}$, $e_{r,2}$, $e_{r,3}$, $...$, $e_{r,n}$''.
A filled example by using the case in Figure~\ref{fig:CAT_overview} is ``relaxing event \texttt{[SEP]} because PersonX wanted \texttt{[SEP]} have fun \texttt{[SEP]} PersonX joins party, go on a holiday, Take a break.''
The commonsense relation within each triple is translated into human-readable text, as shown in Table~\ref{tab:relation_translation}.

\begin{table}[h]
\centering
\small
\begin{tabular}{@{}ll@{}}
\toprule
Relation & Human Readable Text \\ \midrule
xEffect & as a result, PersonX will \\
oEffect & as a result, PersonY or others will \\
xWant & as a result, PersonX want \\
oWant & as a result, PersonY or others want \\
xReact & as a result, PersonX feel \\
oReact & as a result, PersonY or others feel \\
xIntent & because PersonX wanted \\
xNeed & before that, PersonX needed \\
xAttr & PersonX is described as \\ \bottomrule
\end{tabular}
\caption{Textual prompt for commonsense relations~\cite{DBLP:conf/www/FangZWSH21}. Commonsense triple $(h, r, t)$ is translated to human language \textit{“if h, [prompt] t”}.}
\label{tab:relation_translation}
\end{table}

The generative event conceptualization by GPT2 generators uses ``\texttt{[SOS]} $h_o$ \texttt{[SEP]} $i$ \texttt{[GEN]}'' as the input template, where \texttt{[GEN]} indicates the special token for generation.
Commonsense inference modeling uses the same prompt as done by~\citet{DBLP:conf/aaai/HwangBBDSBC21,DBLP:conf/www/FangZWSH21}. 

In addition, we observe that adding special tokens such as \texttt{<c>} and \texttt{</c>} can effectively boost performance.
But adding textual guidelines such as ``is an instance of'' or ``is a concept of'' does not have any positive effect.
The same trend is observed for the bootstrapping prompt, where adding external texts such as ``is also instances of'' or ``can be instantiated to'' will harm the model significantly.

\section{Additional Experiments}
\label{sec:appendix_experiment_detail}

In this section, we present additional details and experiment results for CSKB conceptualization tasks (Appendix~\ref{sec:appendix_CSKB_concept}) and applications, as well as evaluations, of CAT (Appendix~\ref{sec:appendix_application_CAT}) that are not covered in the paper due to limited space.

\subsection{CSKB Conceptualization}
\label{sec:appendix_CSKB_concept}
\subsubsection{Baselines}
\label{sec:appendix_baseline_training_detail}

For supervised learning baselines of both discriminative conceptualization tasks, KG-BERT~\cite{DBLP:journals/corr/abs-1909-03193} is adapted as the skeleton of our baseline models.
For BART, we use the embedding of the end-of-sentence token in the decoder as the representation of the input sequence.
For other models, the embedding of the [CLS] token is used as the representation vector. 
Linear layers are appropriately appended after the encoder model to perform text classification.

For the semi-supervised baselines, we provide additional explanations for different methods: 

\paragraph{UDA.}
In the original paper of UDA~\cite{DBLP:conf/nips/XieDHL020}, two data augmentation methods, back-translation and TF-IDF replacement, are implemented for unsupervised data augmentation. 
We leverage both methods in our conceptualization tasks as two different baselines.
For the triple conceptualization task, we follow the same setting as proposed in PseudoReasoner~\cite{DBLP:journals/corr/abs-2210-07988}.
The back-translation method translates the original corpus from English to French and then translates it back.
Special replacements are taken to avoid the influence of special tokens.  
Meanwhile, the TF-IDF method uses a probability of 0.1 to replace the original corpus according to its TF-IDF score.
For the event conceptualization task, we concatenate the head event and its annotated concept into one new sentence and then feed it into the model. 
For the unlabeled conceptualizations, we enclose the instance and concept with special tokens <c> and </c>, which is the same as our framework, and then use back translation or TF-IDF to generate the augmented data.
The input for triple conceptualization follows a similar way as supervised baselines.
It is observed that these special tokens will not affect the translation significantly as they will be preserved in the translation output.
Last but not least, the model $\theta$ is trained on a mixture of annotated data $x_1$ and augmented data $x_2$ by using the consistency training loss, as shown in Equation~\ref{eq:UDA_loss}.

\begin{equation}
    \label{eq:UDA_loss}
    \begin{multlined}
    J(\theta)  =\mathbb{E}_{x1 \sim P_{L}(x)}[-\log p_{\theta}(y_{1}|x_{1})]+ \\
    \lambda \mathbb{E}_{x2 \sim P_{U}(x)} \mathbb{E}_{\hat{x} \sim q(\hat{x}|x_{2}})[CE(p_{\tilde{\theta}}(y|x_{2})||p_{\theta}(y|\hat{x})]
    \end{multlined}
\end{equation}    

\paragraph{NoisyStudent.}
Noisy Student~\cite{DBLP:conf/cvpr/XieLHL20} is an iterative training method that leverages a teacher-student paradigm.
The teacher model is first trained on annotated data.
It is then asked to make predictions on the unlabeled data as pseudo-labels.
Then, another student model with an equal or larger number of parameters is trained with a mixture of annotated and pseudo-labeled data.
Note that pseudo labels, in numerical values, are directly used as the targeting labels.
The trained student model will serve as a new teacher and re-label the unlabeled data again to yield a better prediction. 
In our implementation, dropout or dynamic model depth is introduced as noise to the model. 
All models $\theta$ are trained with standard cross-entropy loss, as shown in Equation~\ref{eq:BCE_loss}.
We set the dropout probability to 0.5, as it leads to the fastest convergence on our data.
Only one iteration is completed in our experiment, as that's when the student model reaches its best result.

\paragraph{PseudoReasoner.}
PseudoReasoner~\cite{DBLP:journals/corr/abs-2210-07988} is another iterative semi-supervised learning framework that is proposed to tackle Commonsense Knowledge Base Population (CKBP) task~\cite{DBLP:conf/emnlp/FangWCHZSH21,DBLP:journals/corr/abs-2304-10392}.
It leverages a similar teacher-student paradigm and a novel filtering mechanism with the assistance of the student model.
We replaced the generative teacher model with a DeBERTa-v3-large model due to the disastrous performance that GPT2 achieved on both verification tasks.
Similar to CAT, two thresholds, $T^+=0.9$ and $T^-=0.1$, are determined to assign pseudo-labels to unlabeled data based on the prediction of the teacher model.
The rest steps remain the same as described in the original paper.
Similar to NoisyStudent, only one iteration is carried out for PseudoReasoner as the student model converges to the best.

\subsubsection{Settings}
\label{sec:appendix_verification_setting}

We use pretrained language models from the Huggingface Transformers\footnote{\href{https://huggingface.co/docs/transformers/index}{https://huggingface.co/docs/transformers}} Library~\cite{DBLP:conf/emnlp/WolfDSCDMCRLFDS20} to build our framework.
The learning rate for all models is set as 5e-6, and the batch size is 64.
We use an AdamW~\cite{DBLP:conf/iclr/LoshchilovH19} optimizer and evaluate the model every 25 steps.
The max sequence length for the tokenizer is set to 25 and 35 for both discriminative tasks, respectively.
Due to the imbalanced dataset, we evaluate the discriminative models with Area Under Curve (AUC) score~\cite{DBLP:journals/pr/Bradley97}.
Early stopping is used where the best checkpoint is selected when the largest validation AUC is achieved. 
All experiments are repeated three times using different random seeds, and the average performances and standard deviations are reported.
In addition, we set the probability thresholds for both tasks to $T^+=0.9$ and $T^-=0.1$ to determine the pseudo labels. 
The thresholds are roughly derived by observing the overall distribution and quality of data satisfying the respective threshold. 
For the bootstrapping method, we bootstrap $m=9$ additional concepts for event conceptualization verification and $n=2$ additional instances for abstract triple verification. 
Detailed ablation studies are provided in Section~\ref{sec:number_retrieved_alternatives}.
As for the computational infrastructure, the models are trained and evaluated on four NVIDIA RTX3090 (24G) and four NVIDIA 1080Ti (12G) graphical cards. 
The number of parameters for every model is reported in Table~\ref{tab:additional_CAT_performance}.

\subsubsection{Additional Experiment Results}
\label{sec:appendix_verification_result}

The full experiment results for discriminative CSKB conceptualization tasks are reported in Table~\ref{tab:additional_CAT_performance}.
All supervised learning baselines achieve comparable results as reported by~\citet{DBLP:journals/corr/abs-2206-01532}.
Supervised CAT will be discussed later.
The results by semi-supervised CAT are generally consistent with our findings as discussed in Section~\ref{sec:exp_cskb_concept}.
To study the effect of different components and the training regime of CAT, we conduct more detailed ablation studies in Appendix~\ref{sec:ablation_study}.

\subsubsection{Ablation Study}
\label{sec:ablation_study}
In this section, we study the effects of different components in CAT and the training strategy of CAT.
These studies indicate that our framework design and the proposed bootstrapping method play an important role in CSKB conceptualization and are more effective than leveraging unlabeled data with pseudo labels.

\paragraph{Framework Components.}
Our CAT framework consists of three critical components that make CAT different from traditional semi-supervised baselines.
They are denoted as:

\noindent $\bullet$ Bootstrapping:
Assist the training of student models by retrieving alternative conceptualizations and instantiations and bootstrapping them via natural language prompts.
Dropping this component will train student models with the original textual prompts that are also used by the teacher models.

\noindent $\bullet$ CAT Cycle: 
Unite event and triple conceptualization tasks by assigning negative pseudo labels to abstract triples whose conceptualized head is predicted as wrong conceptualization.
Dropping this component will separate the framework into two lines of training, which are training event conceptualization and triple conceptualization models separately.

\noindent $\bullet$ Pseudo-label refinement: 
Refine the pseudo labels with the latest student models and re-train the student models.
Dropping this component will not update any pseudo label and will not re-train the student model.

\begin{table}[h]
    \small
	\centering
        \setlength{\tabcolsep}{4.5pt}
	\begin{tabular}{l|cc}
	\toprule
	\textbf{Models} & \textbf{Event.} & \textbf{Triple.}\\
\midrule
CAT (BERT-base)& \textbf{87.4} & \textbf{76.3}\\
\midrule
$\diamond$ w/o Bootstrapping & 83.1 & 73.0 \\
$\diamond$ w/o CAT Cycle & 86.5 & 75.1 \\
$\diamond$ w/o Pseudo-label Refinement & 87.4 & 76.2 \\
\midrule
\midrule
CAT (DeBERTa-v3-large) & \textbf{89.2} & \textbf{80.0}\\
\midrule
$\diamond$ w/o Bootstrapping & 84.0 & 77.7 \\
$\diamond$ w/o CAT Cycle & 88.1 & 79.0 \\
$\diamond$ w/o Pseudo-label Refinement & 89.1 & 79.7 \\
\bottomrule
\end{tabular}
\caption{Ablation study on three components of CAT. Three components refer to the explanations above. The column \textbf{Event.} indicates test set AUC on the event conceptualization task, and the column \textbf{Triple.} indicates test set AUC on the triple conceptualization task.}
 \label{tab:ablation_study}
\end{table}

We then conduct ablation studies regarding these three components with semi-supervised CAT to prove the effectiveness of our framework design and proposed bootstrapping method.
Each component is removed separately, and the test set performances by student models are reported.
The results are shown in Table~\ref{tab:ablation_study}.
From the results, bootstrapping alternative conceptualization and instantiation leads to the largest performance gain.
Bridging event conceptualization discrimination with triple conceptualization also causes slight improvements.
However, refining the pseudo labels and re-train the student models have barely any effect.
Thus, our bootstrapping method is the most important component within the entire CAT framework and can effectively assist in learning conceptual knowledge.

\paragraph{Supervised CAT.}
We further study training CAT in a supervised learning setting to examine the role of unlabeled data. 
In supervised CAT, no teacher models are trained to provide pseudo labels.
The alternative conceptualizations and instantiations are retrieved directly from the annotated event conceptualization data and bootstrapped later.
Two student models are trained on the bootstrapped data only and evaluated on the same testing set, and the results are reported in Table~\ref{tab:additional_CAT_performance}.
Compared with supervised learning baselines, supervised CAT can achieve a comparable result on the event conceptualization task.
This may be due to the fact that the diversity of concepts drops without considering unlabeled conceptualizations.
Improvements in the triple conceptualization task are more significant, and the results are comparable with semi-supervised CAT.
This indicates that our framework design and bootstrapping method are successful in discriminating high-quality abstract commonsense knowledge, and leveraging a semi-supervised learning paradigm benefits more in event conceptualization discrimination.

\subsection{Application and Evaluation of CAT}
\label{sec:appendix_application_CAT}

\subsubsection{Settings}
\label{sec:appendix_application_CAT_settings}

Pretrained GPT2 models from the Huggingface Transformers Library and training codes\footnote{\href{https://github.com/allenai/comet-atomic-2020}{https://github.com/allenai/comet-atomic-2020}} by~\citet{DBLP:conf/aaai/HwangBBDSBC21} are used as our code base.
The learning rate for all experiments is set to 1e-5, and the batch size is fixed to 64.
We use an Adam~\cite{DBLP:journals/corr/KingmaB14} optimizer and evaluate the model every 20 steps.
The input and output lengths for GPT2 models are fixed at 45 and 55 for the two application and evaluation tasks, respectively.
Such length settings can cover all annotated conceptualizations and triples.
For both generative experiments, we evaluate the generations with BLEU~\cite{DBLP:conf/acl/PapineniRWZ02}, METEOR~\cite{DBLP:conf/wmt/LavieA07}, ROUGE-L~\cite{ROUGE}, and CIDEr~\cite{DBLP:conf/cvpr/VedantamZP15} scores.
However, since an abstract concept usually contains one or two tokens, we only report BLEU1 and BLEU2 scores for the generative event conceptualization task.
Early stopping is also applied where the best checkpoint is selected when the minimum autoregressive LM loss is achieved.
In addition, we notice that the number of triples from the ATOMIC subset is much smaller than abstract triples for the commonsense inference modeling task.
Thus, we upsample the ATOMIC subset by a ratio of 1:2 across all experiments to guarantee a consistent and balanced number of training data.
For generative event conceptualization, the training data is simply a mixture of annotated and pseudo-labeled event conceptualizations without any balancing measure.
All the models are trained and evaluated on four NVIDIA RTX A6000 graphical cards with 48G memory.
The number of parameters is close to the number of parameters in GPT2-XL, which is reported in Table~\ref{tab:additional_CAT_performance}.

\subsubsection{Annotation Settings}
\label{sec:appendix_annotation}

When evaluating the event conceptualization generator, expert annotations are conducted to evaluate concepts that are not presented in the training set.
Crowdsourced platforms such as Amazon Mechanical Turk are not used since experts understand conceptualization better and are more reliable for evaluation.
Subsequently, the authors of this paper are invited to serve as expert annotators.
They are experienced in NLP research and clearly understand the paper's scope.
The annotation guideline is carefully designed.
Each question presents the original head event with the instance highlighted and the corresponding conceptualization candidate to be annotated.
There are also several positive and negative conceptualizations attached as examples.
The authors are well-informed about the instruction and the intended use of their annotations in this paper.
And they all agreed to annotate as part of their contributions.
Moreover, in order to ensure that the expert will not deliberately raise the plausible rate of a certain set of annotation candidates, we randomly shuffle all the data and invite one more expert to cross-validate the annotations.
These measures can ensure that the annotation process is free of ethical concerns and justifiable.

\subsubsection{Additional Experiment Results}
\label{sec:appendix_generation_result}

\begin{figure}[t]
    \centering
    \includegraphics[width=1\linewidth]{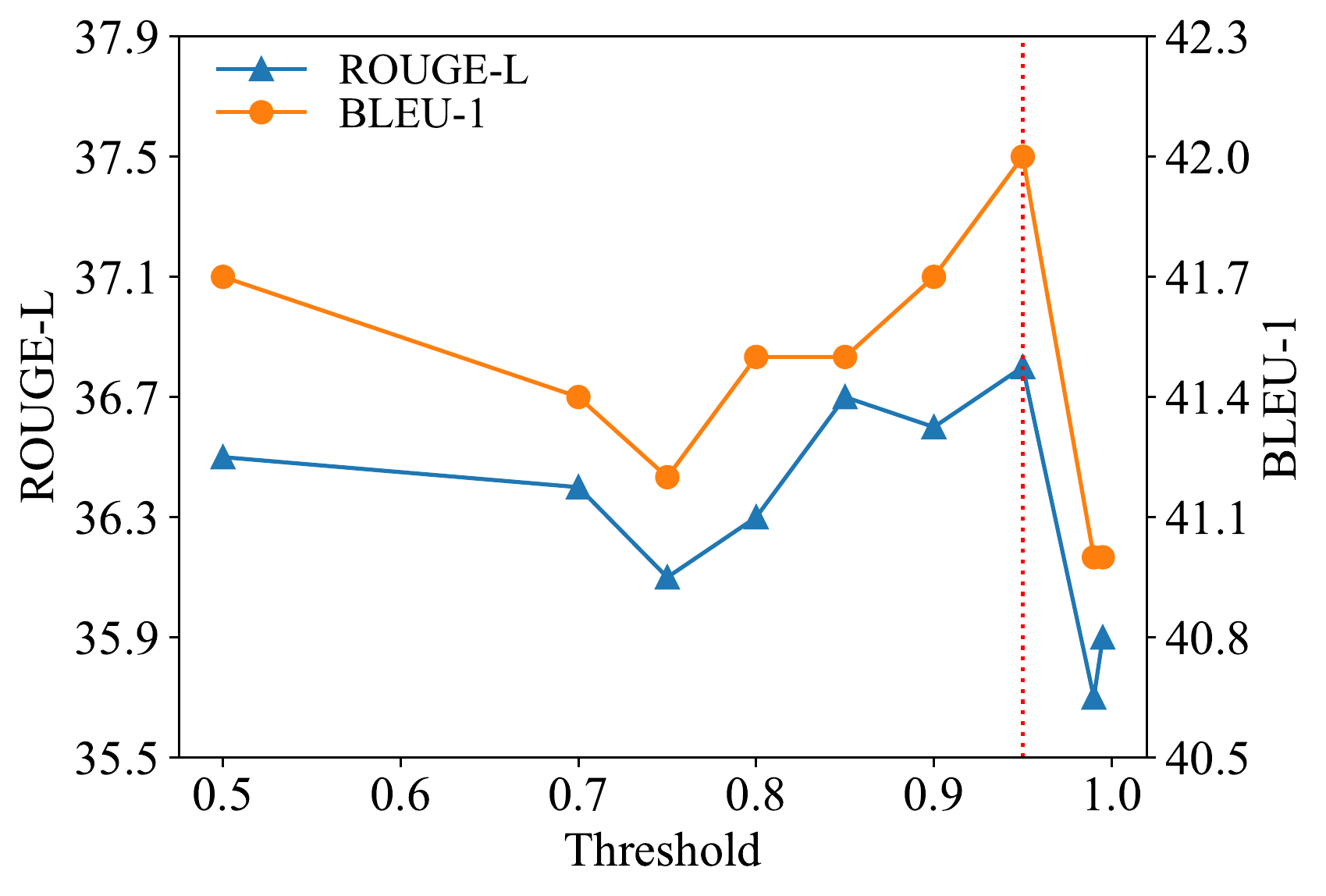}
    \caption{Performance (\%) curve by COMET (GPT2-XL) on commonsense inference generation task with different thresholds for determining positive pseudo labels. 
    Performance with the best threshold of 0.95 is marked as the red dotted line.}
    \label{fig:appendix_COMET_threshold}
\end{figure}

We conduct a more comprehensive study on the commonsense inference generation task by experimenting with the effect of threshold tuning when filtering abstract commonsense knowledge.
Multiple thresholds ranging from 0.5 to 0.995 are experimented with to derive abstract commonsense knowledge of different qualities.
COMET (GPT2-XL) generators are fine-tuned on the ATOMIC subset, augmented by a mixture of annotated and pseudo-labeled abstract triples.
The performance curve according to the threshold is plotted in Figure~\ref{fig:appendix_COMET_threshold}.
Full version results with all metrics are reported in Table~\ref{tab:additional_triple_tail_generation_performance}.
It can be observed that gradually increasing the threshold from 0.75 will lead to better performance, which may be due to the improvement in data quality.
However, increasing the threshold over 0.95 will cause a performance drop.
One possible reason is the amount of pseudo-labeled triples significantly drops with a relatively high threshold, and COMET fails to learn well from annotated triples only.
Using the CAT framework to pseudo-label unlabeled abstract triples leads to better performance than leveraging a RoBERTa-large supervised discriminator to assign pseudo-labels, which also validates the reliability of the triple conceptualization discriminator in CAT.
Also, it is noticeable that training COMET with triples based on our constructed ATOMIC subset is much worse than training with the full ATOMIC dataset.
This indicates that exposing the model with substantial factual commonsense knowledge is still important, and only equipping the model with abstract commonsense knowledge is not enough for commonsense inference modeling.

\section{Computational Cost Analysis}
\label{sec:appendix_computational_cost}

In this section, we compare the number of training data used for both CSKB conceptualization tasks to compare the computational cost across different frameworks and methodologies empirically.
Both annotated and pseudo-labeled data are counted.
The comparison result is presented in Table~\ref{tab:computational_cost}.
All semi-supervised learning methods leverage a significant amount of unlabeled data due to the great scarcity of annotations.
With threshold filterings, PseudoReasoner~\cite{DBLP:journals/corr/abs-2210-07988} and our CAT framework can abandon more than half of pseudo examples with poor quality.
Even though our CAT framework can still outperform PseudoReasoner and achieve the best performance among all methods.
Additionally, there is no notable increase in the number of model parameters as CAT also applies a teacher-student paradigm that is similar to Noisy-Student and PseudoReasoner.
Even compared with the supervised baselines, CAT only doubles the parameters used.
In conclusion, with comparable training data and parameters against other baselines, CAT can achieve much better results and state-of-the-art performances.

\begin{table}[ht]
\small
\centering
\begin{tabular}{@{}l|cc|c@{}}
\toprule
Method & Event. & Triple. & Total \\ 
 \midrule
Supervised Baselines & 107,384 & 65,386 & 172,770 \\
UDA & 412,367 & 4,916,658 & 5,329,025 \\
Noisy-Student & 412,367 & 4,916,658 & 5,329,025 \\
PseudoReasoner & 316,601 & 1,727,865 & 2,044,466 \\ 
\midrule
CAT & 317,507 & 1,595,411 & 1,912,918 \\
\bottomrule
\end{tabular}
\caption{Comparison between the number of training data for discriminative event conceptualization (Event.) and triple conceptualization (Triple.) tasks.}
\label{tab:computational_cost}
\end{table}

\section{Case Studies}
\label{sec:appendix_case_study}
This section contains case studies of the four tasks we studied in this paper, including CSKB conceptualization tasks and applications of CAT.
Throughout these cases, we would like to offer a clearer view of the data, discuss the challenges of the conceptualization task, and provide brief error analyses.

\subsection{CSKB Conceptualization}

\paragraph{Event Conceptualization.}
For discriminative event conceptualization, the case study is shown in Table~\ref{tab:case_study_concept_verification}.
From these cases, it can be observed that several instances $i$ can be identified within one head event $h_o$, and each of them can be conceptualized in multiple ways.
Formally, assume we are conceptualizing $m$ events, each with $n$ instances.
And each instance $i$ concerned can be conceptualized as $p$ concepts.
Each concept takes the majority vote of $q$ annotators to verify.
Subsequently, the number of annotations needed is $O(mnpq)$, which grows significantly if we conceptualize a commonsense knowledge base at scale.
Thus, it is extremely infeasible for practitioners to annotate all of the conceptualizations for verification, which also highlights the importance of a reliable discriminative conceptualization model as CAT acquired.
Semi-supervised learning is also an ideal training strategy, as there is a considerable amount of unlabeled data.

Analyzing the errors made by our discriminator, we observe that models frequently make errors when the instance contains the word ``PersonX,'' which could be caused by the reporting bias~\cite{DBLP:conf/cikm/GordonD13}, as ``PersonX'' is seldom used in normal natural language texts.
Replacing the subjects with commonly used names such as ``Alex, Bob'' may alleviate such a problem.
Additionally, models make errors on some rarely seen concepts, such as ``organ,'' ``cognitive ability,'' and ``side effect.''
Their absence from training data can partially cause this, as CSKB, like ATOMIC, may not cover many instances under those rarely used concepts.

\paragraph{Triple Conceptualization.}
For triple conceptualization discrimination, case studies are shown in Table~\ref{tab:case_study_triple_verification}.
Similar to the analysis above, consider $m$ events with $n$ instances, each instance with $p$ concepts.
Assume that every ATOMIC head event has $t$ relation and tail tuples as its counterpart, and $q$ votes are required from annotators.
The total number of annotations is $O(mnptq)$ for verifying all abstract commonsense triples, which is also huge compared with the total number of original commonsense triples.

The errors are mainly due to the loss of contextualization within the original head events, as conceptualized head events with too high abstractness are likely to omit salient properties.
For example, conceptualizing ``watching a scary movie'' as ``watching movie'' will lose the property ``scary,'' which further leads to a wrong abstract commonsense knowledge if the tail is ``feel scared.''
This also highlights the importance of verifying the plausibility of abstract commonsense knowledge that heavily relies on both the contextualization brought by $r,t$ and the conceptualization of the head event. 
Meanwhile, we observe that the models tend to make a neutral decision (plausibility score close to 0.5) when encountering the situation of conceptualizing an entire event as a concept with high-level abstractness.
Indeed, they are more difficult abstract commonsense knowledge for machines to learn, as a higher level of abstractness leads to more possible instantiations and commonsense inferences.

\begin{table*}[t]
\small
\centering
\begin{tabular}{@{}l|l@{}}
\toprule
Task & Prompt \\ \midrule
Discriminative Event Conceptualization & \parbox[c]{10cm}{\texttt{Given the event} <\textit{event}>, \texttt{can the} <\textit{instance}> \texttt{be conceptualized as} <\textit{concept}>? \texttt{Only answer yes or no without any other words. You are forced to make a decision.}} \\
\midrule
Discriminative Triple Conceptualization & \parbox[c]{10cm}{\texttt{Given a commonsense knowledge triple}, <\textit{head, relation, tail}>, \texttt{is this knowledge plausible or not? Only answer yes or no without any other word. You are forced to make a decision.}} \\
\midrule
Generative Event Conceptualization & \parbox[c]{10cm}{\texttt{Given the event} <\textit{event}>, \texttt{what are possible conceptualizations of} <\textit{instance}>? \texttt{Only list out five short conceptualizations, and do not provide explanations.}} \\ 
\bottomrule
\end{tabular}
\caption{Natural language prompts used to instruct ChatGPT to perform specific tasks. Words in italics and enclosed by brackets indicate inputs replaced by sampled data entries. Restrictive commands are appended at the end to ensure ChatGPT executes the task as intended.}
\label{tab:ChatGPT_prompt}
\vspace{-0.1in}
\end{table*}

\subsection{Appliaction of CAT}

\paragraph{Generative Event Conceptualization.}
The examples are shown in Table~\ref{tab:case_study_concept_generation}.
Generated conceptualizations are generally plausible, given the head event as the context.
Specifically, we observe that neural generators are more sensitive to the instance and its context, as heuristic matching may conceptualize ``sleeping at night'' and ``having trouble sleeping at night'' as ``sleeping''. 
In contrast, neural generators can distinguish these two instances clearly by conceptualizing them as ``sleep'' and ``sleep disorder''.
One potential weakness of neural generators is that the generated conceptualizations lack diversity and novelty~\cite{DBLP:conf/emnlp/DuDLL19,DBLP:conf/acl/WangICR21}, as they tend to be semantically close to the targeting conceptualizations in the training samples.
Nevertheless, it still offers a reliable and simplified approach to performing contextualized conceptualization without tedious matching and human annotations.
Such results also validate the reliability of our discriminative event conceptualization model, as the pseudo-labeled conceptualizations tend to be of high quality.

\paragraph{Commonsense Inference Modeling (COMET).}
Generations from COMET that are only trained on the ATOMIC subset, possibly augmented by abstract commonsense triples, are compared in Table~\ref{tab:case_study_COMET_generation}.
From these generations, we can observe that the abstract commonsense knowledge-aided COMET generator can generate tail events that are most plausible and generalizable compared with the one only trained on ATOMIC.
It generally supports our hypothesis that abstract commonsense knowledge may implicitly help model situational commonsense inference, even without the instantiation step.
In addition, this also validates that our automatically derived abstract knowledge is reliable and helpful, which also proves the reliability of our triple conceptualization discriminator.

\subsection{Conceptualization by Large Language Models}
\label{sec:appendix_ChatGPT}
With the recent advances in Large Language Models (LLMs), such as GPT3.5~\cite{DBLP:conf/nips/BrownMRSKDNSSAA20,DBLP:conf/nips/Ouyang0JAWMZASR22} and ChatGPT~\cite{openai2022chatgpt}, on various NLP tasks~\cite{DBLP:journals/corr/abs-2302-06476,DBLP:journals/corr/abs-2303-16421,DBLP:journals/corr/abs-2304-14827,DBLP:journals/corr/abs-2303-03186}, we also aim to explore ChatGPT's conceptualization ability through case studies.
To do so, we investigate ChatGPT's performance on three conceptualization tasks: discriminative event conceptualization, discriminative triple conceptualization, and generative event conceptualization, all of which are defined in Section~\ref{sec:problem_definition}. 
We randomly sample data entries from AbstractATOMIC and prompt ChatGPT with natural language commands to perform various tasks. 
The prompts used for performing these tasks are listed in Table~\ref{tab:ChatGPT_prompt}.
Specifically, we use OpenAI's API\footnote{The code for the model is \texttt{gpt-3.5-turbo}, and the date of access is May 2023.} to prompt ChatGPT and retrieve its generations.

The case studies for three tasks are presented in Table~\ref{tab:case_study_ChatGPT_concept_verification}, Table~\ref{tab:case_study_ChatGPT_triple_verification}, and Table~\ref{tab:case_study_ChatGPT_concept_generation}, respectively.
These demonstrate ChatGPT's strong conceptualization abilities in both discriminative and generative manners. 
While ChatGPT can accurately determine most event conceptualizations and abstract commonsense knowledge, it still makes some mistakes. 
This highlights the value of training a performant discriminator through CAT, as it can effectively detect incorrect conceptualizations and implausible abstract commonsense knowledge. 
Additionally, ChatGPT tends to conceptualize instances using synonyms~\cite{DBLP:conf/acl/HagiwaraOT06} and hypernyms~\cite{DBLP:conf/emnlp/YuHWSZNS20} and paraphrased or explained terms rather than higher-level concepts. 
This underscores the importance of our event conceptualization generator, which can generate precise, concise event conceptualizations.
In conclusion, our work holds significant value in the realm of commonsense reasoning through conceptualization, particularly in light of the rise of large language models.

\begin{table*}[ht]
\small
\centering
\begin{tabular}{l|l|cc|cc}
\toprule
\multirow{2}{*}{Framework} & \multirow{2}{*}{Backbone PTLM / Method} & \multicolumn{2}{c|}{Event Conceptualization} & \multicolumn{2}{c}{Triple Conceptualization} \\
\cmidrule(l){3-6}
 &  & Validation & Testing & Validation & Testing \\ 
 \midrule
\multirow{14}{*}{\begin{tabular}[c]{@{}l@{}}Supervised\\ Learning\end{tabular}} & BERT-base \scriptsize{\textit{110M}} & 82.4$_{\pm\text{0.05}}$
 & 82.5$_{\pm\text{0.31}}$ & 71.2$_{\pm\text{0.58}}$ & 72.6$_{\pm\text{0.71}}$ \\
& BERT-large \scriptsize{\textit{340M}} & 82.8$_{\pm\text{0.48}}$ & 83.1$_{\pm\text{0.80}}$ & 72.4$_{\pm\text{0.01}}$ & 73.7$_{\pm\text{0.00}}$ \\
& BART-base \scriptsize{\textit{139M}} & 83.8$_{\pm\text{0.28}}$ & 84.4$_{\pm\text{0.32}}$ & 72.0$_{\pm\text{0.09}}$ & 72.6$_{\pm\text{0.15}}$ \\
& BART-large \scriptsize{\textit{406M}} & 85.0$_{\pm\text{0.13}}$ & 85.2$_{\pm\text{0.22}}$ & 74.5$_{\pm\text{0.13}}$ & 76.2$_{\pm\text{0.19}}$ \\
& RoBERTa-base \scriptsize{\textit{110M}} & 84.1$_{\pm\text{0.04}}$ & 84.5$_{\pm\text{0.19}}$ & 72.2$_{\pm\text{0.00}}$ & 74.1$_{\pm\text{0.00}}$ \\
& RoBERTa-large \scriptsize{\textit{340M}} & 85.2$_{\pm\text{0.24}}$ & 85.5$_{\pm\text{0.02}}$ & 75.3$_{\pm\text{0.00}}$ & 76.9$_{\pm\text{0.01}}$ \\
& DeBERTa-v3-base \scriptsize{\textit{214M}} & 85.1$_{\pm\text{0.08}}$ & 85.8$_{\pm\text{0.07}}$ & 73.9$_{\pm\text{0.10}}$ & 75.9$_{\pm\text{0.04}}$ \\
& DeBERTa-v3-large \scriptsize{\textit{435M}} & \underline{85.8$_{\pm\text{0.05}}$} & \underline{86.2$_{\pm\text{0.15}}$} & \underline{76.9$_{\pm\text{0.03}}$} & \underline{78.0$_{\pm\text{0.02}}$} \\
& ELECTRA-base \scriptsize{\textit{110M}} & 85.4$_{\pm\text{0.05}}$ & 85.8$_{\pm\text{0.02}}$ & 74.3$_{\pm\text{0.27}}$ & 76.2$_{\pm\text{0.12}}$ \\
& ELECTRA-large \scriptsize{\textit{340M}} & 84.7$_{\pm\text{0.47}}$ & 85.3$_{\pm\text{0.38}}$ & 75.6$_{\pm\text{0.01}}$ & 77.9$_{\pm\text{0.06}}$ \\
\cmidrule{2-6}
& GPT2-base \scriptsize{\textit{117M}} & 60.0$_{\pm\text{0.06}}$ & 59.1$_{\pm\text{0.14}}$ & 52.8$_{\pm\text{0.14}}$ & 55.9$_{\pm\text{0.11}}$ \\
& GPT2-medium \scriptsize{\textit{345M}} & 61.2$_{\pm\text{0.11}}$ & 60.3$_{\pm\text{0.08}}$ & 54.6$_{\pm\text{0.17}}$ & 57.4$_{\pm\text{0.09}}$ \\
& GPT2-large \scriptsize{\textit{774M}} & 64.1$_{\pm\text{0.05}}$ & 62.7$_{\pm\text{0.08}}$ & 60.5$_{\pm\text{0.11}}$ & 59.8$_{\pm\text{0.06}}$ \\
& GPT2-XL \scriptsize{\textit{1558M}} & 64.2$_{\pm\text{0.19}}$ & 63.6$_{\pm\text{0.22}}$ & 62.2$_{\pm\text{0.08}}$ & 61.5$_{\pm\text{0.10}}$ \\
\midrule
\multirow{5}{*}{\begin{tabular}[c]{@{}l@{}}Semi-Supervised\\ Learning\end{tabular}} & UDA (TF-IDF) &83.6$_{\pm\text{0.29}}$ &83.6$_{\pm\text{0.24}}$ & 75.8$_{\pm\text{1.26}}$ & 76.8$_{\pm\text{1.34}}$ \\ 
& UDA (back-trans.) &83.4$_{\pm\text{0.27}}$ & 83.6$_{\pm\text{0.24}}$ & 75.8$_{\pm\text{1.25}}$ & 76.8$_{\pm\text{1.34}}$ \\
& Noisy-Student & 86.4$_{\pm\text{0.05}}$ & 86.5$_{\pm\text{0.09}}$ & 75.4$_{\pm\text{0.64}}$ & 76.7$_{\pm\text{0.59}}$ \\
& PseudoReasoner (BERT-base) & 83.3$_{\pm\text{0.11}}$ & 84.0$_{\pm\text{0.24}}$ & 73.0$_{\pm\text{0.14}}$ & 74.1$_{\pm\text{0.33}}$ \\
& PseudoReasoner (RoBERTa-large) & \underline{86.6$_{\pm\text{0.25}}$} & \underline{86.7$_{\pm\text{0.33}}$} & \underline{76.3$_{\pm\text{0.12}}$} & \underline{77.2$_{\pm\text{0.21}}$} \\
\midrule
\multirow{10}{*}{\begin{tabular}[c]{@{}l@{}}\textbf{CAT}\\ \textbf{\textit{(Supervised)}}\end{tabular}} & BERT-base \scriptsize{\textit{110M}} & 83.9$_{\pm\text{0.42}}$ & 84.5$_{\pm\text{0.43}}$ & 73.4$_{\pm\text{0.32}}$ & 73.3$_{\pm\text{0.23}}$ \\
& BERT-large \scriptsize{\textit{340M}} & 82.8$_{\pm\text{0.48}}$ & 83.1$_{\pm\text{0.80}}$ & 72.4$_{\pm\text{0.01}}$ & 73.7$_{\pm\text{0.00}}$ \\
& BART-base \scriptsize{\textit{139M}} & 84.9$_{\pm\text{0.05}}$ & 85.4$_{\pm\text{0.08}}$ & 75.2$_{\pm\text{0.06}}$ & 76.9$_{\pm\text{0.21}}$ \\
& BART-large \scriptsize{\textit{406M}} & 86.2$_{\pm\text{0.05}}$ & 86.0$_{\pm\text{0.06}}$ & 76.8$_{\pm\text{0.21}}$ & 78.7$_{\pm\text{0.31}}$ \\
& RoBERTa-base \scriptsize{\textit{110M}} & 85.5$_{\pm\text{0.06}}$ & 86.0$_{\pm\text{0.06}}$ & 76.6$_{\pm\text{0.12}}$ & 77.2$_{\pm\text{0.18}}$ \\
& RoBERTa-large \scriptsize{\textit{340M}} & 86.2$_{\pm\text{0.31}}$ & 86.2$_{\pm\text{0.31}}$ & 77.7$_{\pm\text{0.19}}$ & 78.5$_{\pm\text{0.28}}$ \\
& DeBERTa-v3-base \scriptsize{\textit{214M}} & 85.8$_{\pm\text{0.15}}$ & 86.2$_{\pm\text{0.07}}$ & 76.8$_{\pm\text{0.28}}$ & 79.0$_{\pm\text{0.20}}$ \\
& DeBERTa-v3-large \scriptsize{\textit{435M}} & \underline{86.3$_{\pm\text{0.11}}$} & \underline{86.7$_{\pm\text{0.08}}$} & \underline{78.4$_{\pm\text{0.20}}$} & \underline{79.5$_{\pm\text{0.18}}$} \\
& ELECTRA-base \scriptsize{\textit{110M}} & 85.5$_{\pm\text{0.12}}$ & 85.7$_{\pm\text{0.08}}$ & 76.7$_{\pm\text{0.05}}$ & 77.3$_{\pm\text{0.16}}$ \\
& ELECTRA-large \scriptsize{\textit{340M}} & 86.2$_{\pm\text{0.66}}$ & 86.0$_{\pm\text{0.62}}$ & 77.8$_{\pm\text{0.11}}$ & 78.5$_{\pm\text{0.09}}$ \\
\midrule
\multirow{10}{*}{\begin{tabular}[c]{@{}l@{}}\textbf{CAT}\\ \textbf{\textit{(Semi-Supervised)}}\end{tabular}} & BERT-base \scriptsize{\textit{110M}} & 87.1$_{\pm\text{0.06}}$ & 87.4$_{\pm\text{0.11}}$ & 74.3$_{\pm\text{0.26}}$ & 76.3$_{\pm\text{0.38}}$ \\
& BERT-large \scriptsize{\textit{340M}} & 87.7$_{\pm\text{0.16}}$ & 88.0$_{\pm\text{0.19}}$ & 75.8$_{\pm\text{0.23}}$ & 77.8$_{\pm\text{0.36}}$ \\
& BART-base \scriptsize{\textit{139M}} & 88.2$_{\pm\text{0.09}}$ & 88.2$_{\pm\text{0.09}}$ & 75.7$_{\pm\text{0.09}}$ & 78.0$_{\pm\text{0.14}}$ \\
& BART-large \scriptsize{\textit{406M}} & 88.6$_{\pm\text{0.07}}$ & 88.7$_{\pm\text{0.10}}$& 77.2$_{\pm\text{0.12}}$ & 79.0$_{\pm\text{0.14}}$ \\
& RoBERTa-base \scriptsize{\textit{110M}} & 88.4$_{\pm\text{0.12}}$ & 88.3$_{\pm\text{0.08}}$ & 76.9$_{\pm\text{0.16}}$ & 78.0$_{\pm\text{0.19}}$ \\
& RoBERTa-large \scriptsize{\textit{340M}} & 89.0$_{\pm\text{0.15}}$ & 88.8$_{\pm\text{0.20}}$ & 78.2$_{\pm\text{0.08}}$ & 79.4$_{\pm\text{0.14}}$ \\
& DeBERTa-v3-base \scriptsize{\textit{214M}} & 88.8$_{\pm\text{0.12}}$ & 88.9$_{\pm\text{0.08}}$ & 77.5$_{\pm\text{0.10}}$ & 79.9$_{\pm\text{0.07}}$ \\
& DeBERTa-v3-large \scriptsize{\textit{435M}} & \textbf{\underline{89.1$_{\pm\text{0.05}}$}} & \textbf{\underline{89.2$_{\pm\text{0.14}}$}} & \textbf{\underline{78.7$_{\pm\text{0.16}}$}} & \textbf{\underline{80.0$_{\pm\text{0.33}}$}} \\
& ELECTRA-base \scriptsize{\textit{110M}} & 88.7$_{\pm\text{0.10}}$ & 88.9$_{\pm\text{0.10}}$ & 74.9$_{\pm\text{0.15}}$ & 75.5$_{\pm\text{0.40}}$ \\
& ELECTRA-large \scriptsize{\textit{340M}} & 88.6$_{\pm\text{0.77}}$ & 88.5$_{\pm\text{0.70}}$ & 74.9$_{\pm\text{0.15}}$ & 75.5$_{\pm\text{0.40}}$ \\
\bottomrule
\end{tabular}
\caption{
Full experiment results (\%) by our CAT framework on the discriminative event conceptualization and triple conceptualization tasks. 
We report the average AUC score and standard deviation across experiments with three random seeds.
The best performances within each framework are underlined, and the best among all models are bold-faced.
All supervised baselines are comparable with experiment results by~\citet{DBLP:journals/corr/abs-2206-01532}.
}
\label{tab:additional_CAT_performance}
\vspace{-0.1in}
\end{table*}

\begin{table*}[h]
\small
\centering
\begin{tabular}{@{}l|c|c|cc@{}}
\toprule
Head Event & Instance & Concept & Label & Pred. \\ 
\midrule
\multirow{8}{*}{\begin{tabular}[c]{@{}l@{}}PersonX accepts\\ the invitation\end{tabular}} & the invitation & personal communication & \checkmark & \checkmark \\
 & the invitation & party idea & $\times$ & \checkmark \\
 & the invitation & friendly approach & \checkmark & \checkmark \\
 & the invitation & item & $\times$ & \checkmark \\
 & PersonX accepts the invitation & acceptance & \checkmark & \checkmark \\
 & PersonX accepts the invitation & approach & $\times$ & $\times$ \\
 & PersonX accepts the invitation & psychological treatment & $\times$ & $\times$\\
 & PersonX accepts the invitation & personal communication & \checkmark & \checkmark \\
 \midrule
\multirow{8}{*}{\begin{tabular}[c]{@{}l@{}}PersonX makes oatmeal \\ for breakfast\end{tabular}} & oatmeal & ingredient & $\times$ & \checkmark \\
 & oatmeal & cereal & \checkmark & \checkmark \\
 & oatmeal & grain food & \checkmark & \checkmark \\
 & breakfast & service & $\times$ & $\times$ \\
 & breakfast & meal & \checkmark & \checkmark \\
 & PersonX makes oatmeal for breakfast & hands-on activity & \checkmark & \checkmark \\
 & PersonX makes oatmeal for breakfast & extended school activity & $\times$ & \checkmark \\
 & PersonX makes oatmeal for breakfast & cooking & \checkmark & \checkmark \\
\bottomrule
\end{tabular}
\caption{Case study of ChatGPT's discriminative event conceptualizations. Label refers to annotation result and Pred. stands for prediction by ChatGPT.}
\label{tab:case_study_ChatGPT_concept_verification}
\vspace{-0.1in}
\end{table*}

\begin{table*}[h]
\small
\centering
\begin{tabular}{l|c|c|cc}
\toprule
Conceptualized Head Event & Relation & Tail Event & Label & Pred. \\ 
\midrule
\multirow{8}{*}{\underline{medical check}} & xEffect & to be brave & \checkmark & $\times$ \\
 & xWant & take medicine & \checkmark & \checkmark \\
 & xWant & leave the hotel & $\times$ & \checkmark \\ 
 & xWant & to drive home & $\times$ & \checkmark \\ 
 & xAttr & cautious & \checkmark & \checkmark \\ 
 & xWant & go to rest & \checkmark & \checkmark \\ 
 & xAttr & diseased & \checkmark & $\times$ \\ 
 & xNeed & get injured & \checkmark & $\times$ \\ 
 \midrule
\multirow{8}{*}{\underline{watching movie}} & xEffect & laugh & \checkmark & $\times$ \\
 & xWant & to be entertained & \checkmark & \checkmark \\
 & xNeed & to go to video store & \checkmark & $\times$ \\ 
 & xIntent & entertain themselves & \checkmark & \checkmark \\ 
 & xWant & to put movie in DVD player & \checkmark & \checkmark \\ 
 & xAttr & satisfied & \checkmark & $\times$ \\ 
 & xReact & scared & \checkmark & \checkmark \\ 
 & xNeed & rent it on Netflix & \checkmark & \checkmark \\ 
 \bottomrule
\end{tabular}
\caption{Case study of ChatGPT's discriminative triple conceptualizations. Underlined words in the head event refer to conceptualizations. Label refers to annotation result and Pred. stands for prediction by ChatGPT.}
\label{tab:case_study_ChatGPT_triple_verification}
\vspace{-0.1in}
\end{table*}

\begin{table*}[h]
\small
\centering
{\def\arraystretch{1.6}
\begin{tabularx}{\textwidth}{>{\raggedright}p{3.5cm}>{\RaggedRight\arraybackslash}p{3cm}p{3cm}p{5cm}} \toprule
Event & Target Conceptualizations & CAT's Generations & ChatGPT's Generations \\ 
\midrule
PersonX is having trouble sleeping at \underline{night} & time, night, nonwork time, shift & night, evening, time, late, darknight & Dark sleep time, nocturnal period, resting hours, nighttime sleeplessness, bedtime wakefulness \\
PersonX is having trouble \underline{sleeping at night} & relaxation, sleeping, resting & rest, sleep, resting, daily routine, sleeping & Nightly slumber attempt, dark rest struggle, sleeplessness after dark, nocturnal insomnia, bedtime wakefulness \\
\underline{PersonX is having trouble} \underline{sleeping at night} & sleeping, disorder, sleep problem, trouble, insomnia, sleep disorder & sleep disorder, problem, sleep disturbance, difficulty, trouble sleeping & Nighttime sleep difficulty, dark restlessness problem, nocturnal insomnia struggle, bedtime wakefulness issue, sleeplessness after dark challenge \\
\bottomrule
\end{tabularx}
}
\caption{Case study of ChatGPT's generative event conceptualizations. The instance candidate in each event is underlined. Target conceptualizations are positive conceptualizations extracted from AbstractATOMIC, including the annotated conceptualizations and ones that are positively pseudo-labeled by our framework.}
\label{tab:case_study_ChatGPT_concept_generation}
\vspace{-0.1in}
\end{table*}

\begin{table*}[h]
\small
\centering
\begin{tabular}{@{}l|c|c|cc@{}}
\toprule
Head Event & Instance & Concept & Label & Pred. \\ 
\midrule
\multirow{8}{*}{\begin{tabular}[c]{@{}l@{}}PersonX is having trouble\\ sleeping at night\end{tabular}} & night & nonwork time & \checkmark & \checkmark \\
 & night & night & \checkmark & \checkmark \\
 & sleeping at night & lifestyle factor & \checkmark & $\times$ \\
 & sleeping at night & basic need & \checkmark & \checkmark \\
 & trouble sleeping at night & board game & $\times$ & $\times$ \\
 & trouble sleeping at night & problem & \checkmark & \checkmark \\
 & PersonX is having trouble sleeping at night & variable & $\times$ & $\times$ \\
 & PersonX is having trouble sleeping at night & personal characteristic & \checkmark & \checkmark \\
 \midrule
\multirow{8}{*}{\begin{tabular}[c]{@{}l@{}}PersonX is nervous about\\ making friends\end{tabular}} & friends & person & \checkmark & \checkmark \\
 & friends & support person & \checkmark & \checkmark \\
 & making friends & relationship & \checkmark & \checkmark \\
 & making friends & social activity & \checkmark & \checkmark \\
 & nervous about making friends & organ & $\times$ & \checkmark \\
 & nervous about making friends & side effect & $\times$ & \checkmark \\
 & PersonX is nervous about making friends & emotion & \checkmark & \checkmark \\
 & PersonX is nervous about making friends & nervous disorder & \checkmark & \checkmark \\
\midrule
\multirow{8}{*}{\begin{tabular}[c]{@{}l@{}}PersonX wants to learn how\\ to play the piano\end{tabular}} & the piano & instrument & \checkmark & \checkmark \\
 & the piano & western instrument & \checkmark & \checkmark \\
 & how to play the piano & musical activity & \checkmark & \checkmark \\
 & how to play the piano & play & \checkmark & $\times$ \\
 & to learn how to play the piano & button & $\times$ & $\times$ \\
 & to learn how to play the piano & learning activity & \checkmark & \checkmark \\
 & PersonX wants to learn how to play the piano & cultural event & $\times$ & $\times$ \\
 & PersonX wants to learn how to play the piano & cognitive ability & \checkmark & $\times$ \\
 \midrule
\multirow{8}{*}{\begin{tabular}[c]{@{}l@{}}PersonX puts PersonX's pants \\ on PersonX's leg at a time\end{tabular}} & PersonX's pants & pant & \checkmark & $\times$ \\
 & PersonX's pants & clothing & \checkmark & \checkmark \\
 & PersonX's leg & leg & \checkmark & $\times$ \\
 & PersonX's leg & limb & \checkmark & $\times$ \\
 & a time & resource & $\times$ & $\times$ \\
 & a time & time & \checkmark & \checkmark \\
 & PersonX puts PersonX's pants on PersonX's leg & dressing & \checkmark & \checkmark \\
 & PersonX puts PersonX's pants on PersonX's leg & action & $\times$ & $\times$ \\
\bottomrule
\end{tabular}
\caption{Case study of CAT's discriminative event conceptualizations. A head event can be conceptualized in multiple ways, as shown in the table. Label refers to annotation result and Pred. stands for prediction by our framework.}
\label{tab:case_study_concept_verification}
\vspace{-0.1in}
\end{table*}

\begin{table*}[h]
\small
\centering
{\def\arraystretch{1.6}
\begin{tabularx}{\textwidth}{>{\raggedright}p{4cm}>{\RaggedRight\arraybackslash}p{5.3cm}p{5.3cm}} \toprule
Event & Target Conceptualizations & Generated Conceptualizations \\ 
\midrule
PersonX is having trouble sleeping at \underline{night} & time, night, nonwork time, shift & night, evening, time, late, darknight \\
PersonX is having trouble \underline{sleeping at night} & relaxation, sleeping, resting & rest, sleep, resting, daily routine, sleeping \\
\underline{PersonX is having trouble} \underline{sleeping at night} & sleeping, disorder, sleep problem, trouble, insomnia, sleep disorder & sleep disorder, problem, sleep disturbance, difficulty, trouble sleeping \\
\midrule
PersonX gets \underline{great grades} in school & accomplishment, result, grades, good performance, achievement & achievement, grades, good grade, academic excellence, grade \\
\underline{PersonX asks what was wrong} & problems, concern, seeking information, questioning, query, communication & query, question, asking, communication, inquiry \\
\underline{PersonX needs new shoes} & necessity, product, personal item, item, clothing, shoes & requirement, item, need, necessity, needs \\
\underline{PersonX is failing math} & negative experience, negative issue, problem, poor performance & difficulty, poor performance, problem, academic failure, math problem \\
\bottomrule
\end{tabularx}
}
\caption{Case study of CAT's generative event conceptualizations. The instance candidate in each event is underlined. Target conceptualizations are positive conceptualizations extracted from AbstractATOMIC, including the annotated conceptualizations and ones that are positively pseudo-labeled by our framework.}
\label{tab:case_study_concept_generation}
\vspace{-0.1in}
\end{table*}

\begin{table*}[h]
\small
\centering
\begin{tabular}{l|c|c|cc}
\toprule
Conceptualized Head Event & Relation & Tail Event & Label & Pred. \\ 
\midrule
\multirow{8}{*}{PersonX gets \underline{nailcare service}} & xAttr & rich & \checkmark & \checkmark \\
 & xAttr & skillful & $\times$ & \checkmark \\
 & xIntent & look pretty & \checkmark & \checkmark \\
 & xNeed & book an appointment & \checkmark & \checkmark \\
 & xEffect & show off & \checkmark & \checkmark \\
 & xReact & excited & \checkmark & \checkmark \\
 & oWant & to tell her they like them & \checkmark & \checkmark \\
 & xWant & to go home & \checkmark & \checkmark \\
 \midrule
\multirow{8}{*}{\underline{watching movie}} & xEffect & laugh & \checkmark & \checkmark \\
 & xWant & to be entertained & \checkmark & \checkmark \\
 & xNeed & to go to video store & \checkmark & $\times$ \\ 
 & xIntent & entertain themselves & \checkmark & \checkmark \\ 
 & xWant & to put movie in DVD player & \checkmark & \checkmark \\ 
 & xAttr & satisfied & \checkmark & \checkmark \\ 
 & xReact & scared & \checkmark & $\times$ \\ 
 & xNeed & rent it on Netflix & \checkmark & \checkmark \\ 
 \midrule
\multirow{8}{*}{\underline{medical check}} & xEffect & to be brave & \checkmark & \checkmark \\
 & xWant & take medicine & \checkmark & \checkmark \\
 & xWant & leave the hotel & $\times$ & \checkmark \\ 
 & xWant & to drive home & $\times$ & \checkmark \\ 
 & xAttr & cautious & \checkmark & \checkmark \\ 
 & xWant & go to rest & \checkmark & \checkmark \\ 
 & xAttr & diseased & \checkmark & \checkmark \\ 
 & xNeed & get injured & \checkmark & \checkmark \\ 
 \bottomrule
\end{tabular}
\caption{Case study of CAT's discriminative triple conceptualizations. The abstract concept within each conceptualized head event is underlined. Label refers to annotation result and Pred. stands for prediction by our framework.}
\label{tab:case_study_triple_verification}
\vspace{-0.1in}
\end{table*}

\begin{table*}[h]
\small
\centering
\begin{tabular}{@{}l|c|c|c@{}}
\toprule
Head & Relation & Source & Tail \\ 
\midrule
\multirow{3}{*}{PersonX washes PersonY's car} & \multirow{3}{*}{oWant} & ATOMIC & to tip PersonX \\
 &  & COMET$_\text{ATOMIC}$ & to wash their car \\
 &  & COMET$_\text{CAT}$ & to thank PersonX \\ 
 \midrule
\multirow{3}{*}{PersonX meets PersonX's standards} & \multirow{3}{*}{xNeed} & ATOMIC & to practice \\
 &  & COMET$_\text{ATOMIC}$ & to study \\
 &  & COMET$_\text{CAT}$ & to practice hard \\ 
 \midrule
\multirow{3}{*}{PersonX stretches out PersonX's hand} & \multirow{3}{*}{xWant} & ATOMIC & to give PersonY something \\
 &  & COMET$_\text{ATOMIC}$ & to touch \\
 &  & COMET$_\text{CAT}$ & to grab something for PersonY \\ 
 \midrule
\multirow{3}{*}{PersonX learns how to bake a cake} & \multirow{3}{*}{xAttr} & ATOMIC & interested \\
 &  & COMET$_\text{ATOMIC}$ & curious \\
 &  & COMET$_\text{CAT}$ & skilled \\ 
 \midrule
\multirow{3}{*}{PersonX fails PersonX's class} & \multirow{3}{*}{xWant} & ATOMIC & to retake the class \\
 &  & COMET$_\text{ATOMIC}$ & to study hard \\
 &  & COMET$_\text{CAT}$ & to try again in the class \\ 
 \midrule
\multirow{3}{*}{PersonX buys dog food} & \multirow{3}{*}{xEffect} & ATOMIC & X gets receipt \\
 &  & COMET$_\text{ATOMIC}$ & loses weight \\
 &  & COMET$_\text{CAT}$ & gets a receipt \\ 
 \midrule
\multirow{3}{*}{PersonX hits by lightning} & \multirow{3}{*}{xEffect} & ATOMIC & has hair burned \\
 &  & COMET$_\text{ATOMIC}$ & gets electrocuted \\
 &  & COMET$_\text{CAT}$ & screams in pain \\ 
  \midrule
\multirow{3}{*}{PersonX forgets my wallet} & \multirow{3}{*}{xEffect} & ATOMIC & is chastised \\
 &  & COMET$_\text{ATOMIC}$ & gets robbed \\
 &  & COMET$_\text{CAT}$ & thinks about it \\ 
  \midrule
\multirow{3}{*}{PersonX realizes something} & \multirow{3}{*}{xWant} & ATOMIC & make a plan \\
 &  & COMET$_\text{ATOMIC}$ & to solve the problem \\
 &  & COMET$_\text{CAT}$ & to do something about it \\ 
\bottomrule
\end{tabular}
\caption{
Case study of commonsense inference generation (COMET). 
Examples are selected from the original ATOMIC testing set. 
ATOMIC refers to the target tail in the original ATOMIC. 
COMET$_\text{ATOMIC}$ and COMET$_\text{CAT}$ stand for generations by COMET trained on an ATOMIC subset or aided with abstract knowledge derived by CAT.}
\label{tab:case_study_COMET_generation}
\vspace{-0.1in}
\end{table*}

\begin{table*}[ht]
\small
\centering
\begin{tabular}{@{}l|cc|cc|cc|cc|cc|cc|cc@{}}
\toprule
\multirow{2}{*}{Training Data} & \multicolumn{2}{c|}{BLEU-1} & \multicolumn{2}{c|}{BLEU-2} & \multicolumn{2}{c|}{BLEU-3} & \multicolumn{2}{c|}{BLEU-4} & \multicolumn{2}{c|}{METEOR} & \multicolumn{2}{c|}{ROUGE-L} & \multicolumn{2}{c}{CIDEr} \\ \cmidrule(l){2-15}
 & Dev & Test & Dev & Test & Dev & Test & Dev & Test & Dev & Test & Dev & Test & Dev & Test \\ 
\midrule
Zero-Shot & 5.42 & 4.89 & 1.84 & 1.51 & 0.65 & 0.52 & 0.26 & 0.21 & 6.50 & 5.70 & 6.40 & 5.90 & 1.60 & 1.20 \\
ATOMIC (subset) & 38.1 & 38.1 & 25.4 & 25.7 & 18.7 & 18.8 & 15.5 & 15.7 & 14.9 & 14.9 & 33.0 & 33.2 & 27.6 & 27.8 \\
\midrule
$+D^l_t$ & 38.1 & 38.5 & 24.8 & 25.5 & 17.8 & 18.4 & 14.7 & 15.2 & 15.3 & 15.6 & 33.1 & 33.7 & 26.8 & 27.3 \\
\hspace{3mm}$+\text{Finetune}$ & 38.6 & 39.0 & 25.8 & 26.6 & 18.9 & 19.7 & 15.7 & 16.4 & 15.1 & 15.4 & 33.6 & 34.4 & 28.8 & 30.0 \\
$+D^{u}_{\text{Abs.ATM.}}$ & 40.0 & 40.3 & 27.1 & 27.8 & 20.0 & 20.8 & 16.5 & 17.5 & 16.1 & 16.3 & 35.3 & 35.7 & 31.6 & 31.7 \\
\hspace{3mm}$+\text{Finetune}$ & 40.1 & 40.5 & 27.1 & 27.8 & 20.1 & 20.8 & 16.7 & 17.4 & 16.2 & 16.4 & 35.4 & 35.9 & 31.8 & 31.7 \\
$+D^l_t+D^{u}_{\text{Abs.ATM.}}$ & 40.2 & 40.6 & 26.2 & 27.4 & 19.0 & 20.4 & 15.1 & 16.8 & 16.3 & 16.5 & 35.0 & 35.4 & 31.0 & 31.3 \\
\hspace{3mm}$+\text{Finetune}$ & 40.0 & 40.4 & 26.0 & 26.9 & 18.7 & 19.7 & 15.0 & 16.1 & 16.3 & 16.4 & 35.0 & 35.4 & 30.3 & 30.7 \\
\midrule
$+D^u_{0.995}$ & 39.7 & 39.8 & 26.5 & 26.8 & 19.5 & 19.8 & 15.6 & 16.1 & 15.8 & 15.8 & 35.0 & 34.9 & 30.8 & 30.7 \\
\hspace{3mm}$+\text{Finetune}$ & 41.0 & 41.0 & 27.1 & 27.5 & 20.0 & 20.2 & 16.1 & 16.3 & 16.7 & 16.6 & 36.0 & 35.9 & 31.9 & 31.7 \\
$+D^u_{0.99}$ & 39.5 & 39.9 & 26.1 & 27.0 & 19.3 & 20.0 & 15.9 & 16.6 & 15.7 & 15.9 & 34.7 & 34.8 & 30.6 & 30.8 \\
\hspace{3mm}$+\text{Finetune}$ & 40.8 & 41.0 & 27.0 & 27.6 & 20.0 & 20.5 & 16.2 & 16.9 & 16.7 & 16.6 & 35.8 & 35.7 & 31.9 & 31.6 \\
\rowcolor{Gray}
$+D^u_{0.95}$ & 41.2 & 41.9 & 28.1 & 29.0 & 20.7 & 21.5 & 16.5 & 17.8 & 16.6 & 16.9 & 35.9 & 36.5 & 33.4 & 33.7 \\
\rowcolor{Gray}
\hspace{3mm}$+\text{Finetune}$ & 41.1 & 42.0 & 28.0 & 29.0 & 20.4 & 21.5 & 16.4 & 17.6 & 16.6 & 17.0 & 36.0 & 36.8 & 33.2 & 33.8 \\
$+D^u_{0.90}$ & 41.6 & 41.6 & 28.1 & 28.5 & 20.9 & 21.5 & 17.1 & 17.7 & 16.9 & 16.8 & 36.7 & 36.4 & 33.4 & 33.1 \\
\hspace{3mm}$+\text{Finetune}$ & 41.8 & 41.7 & 28.3 & 28.5 & 21.0 & 21.4 & 17.0 & 17.5 & 17.0 & 17.0 & 36.7 & 36.6 & 33.4 & 33.1 \\
$+D^u_{0.85}$ & 41.3 & 41.4 & 27.8 & 28.1 & 20.7 & 21.1 & 16.8 & 17.6 & 16.7 & 16.8 & 36.3 & 36.6 & 32.6 & 32.9 \\
\hspace{3mm}$+\text{Finetune}$ & 41.5 & 41.5 & 27.9 & 28.2 & 20.6 & 21.1 & 16.8 & 17.5 & 16.8 & 16.9 & 36.3 & 36.7 & 32.6 & 33.0 \\
$+D^u_{0.80}$ & 41.6 & 41.6 & 27.3 & 28.0 & 20.1 & 20.7 & 16.3 & 17.0 & 17.0 & 16.9 & 36.6 & 36.4 & 33.0 & 32.6 \\
\hspace{3mm}$+\text{Finetune}$ & 41.6 & 41.5 & 27.5 & 27.9 & 20.2 & 20.6 & 16.3 & 16.8 & 17.0 & 16.9 & 36.6 & 36.3 & 33.0 & 32.3 \\
$+D^u_{0.75}$ & 40.6 & 40.8 & 27.1 & 28.0 & 19.9 & 20.9 & 16.2 & 17.2 & 16.4 & 16.6 & 35.5 & 35.7 & 31.6 & 32.1 \\
\hspace{3mm}$+\text{Finetune}$ & 40.9 & 41.2 & 27.2 & 28.1 & 19.9 & 21.0 & 16.2 & 17.0 & 16.6 & 16.9 & 35.7 & 36.1 & 31.8 & 32.7 \\
$+D^u_{0.70}$ & 40.6 & 40.9 & 27.1 & 27.8 & 19.9 & 20.7 & 16.6 & 17.2 & 16.4 & 16.6 & 35.6 & 36.1 & 31.6 & 32.4 \\
\hspace{3mm}$+\text{Finetune}$ & 41.4 & 41.4 & 27.5 & 28.1 & 20.1 & 21.0 & 16.4 & 17.4 & 16.9 & 16.9 & 36.2 & 36.4 & 32.5 & 33.0 \\
$+D^u_{0.50}$ & 41.1 & 41.5 & 27.3 & 28.2 & 20.4 & 21.2 & 16.7 & 17.6 & 16.7 & 16.7 & 35.8 & 36.1 & 32.4 & 32.8 \\
\hspace{3mm}$+\text{Finetune}$ & 41.5 & 41.7 & 27.7 & 28.5 & 20.7 & 21.4 & 17.0 & 17.8 & 16.9 & 16.9 & 36.3 & 36.5 & 32.7 & 33.1 \\
\midrule
$+D^l_t+D^u_{0.995}$ & 39.4 & 39.3 & 26.1 & 26.4 & 19.2 & 19.5 & 15.5 & 15.8 & 15.7 & 15.5 & 33.9 & 33.8 & 29.8 & 29.2 \\
\hspace{3mm}$+\text{Finetune}$ & 39.7 & 40.0 & 26.7 & 27.5 & 19.5 & 20.3 & 15.8 & 16.6 & 15.7 & 15.7 & 34.7 & 34.9 & 30.6 & 30.9 \\
$+D^l_t+D^u_{0.99}$ & 39.4 & 39.7 & 25.7 & 26.5 & 18.6 & 19.5 & 15.2 & 16.5 & 15.8 & 15.9 & 34.6 & 35.0 & 29.7 & 30.2 \\
\hspace{3mm}$+\text{Finetune}$ & 39.7 & 40.4 & 26.6 & 27.6 & 19.6 & 20.5 & 16.0 & 16.8 & 15.7 & 16.1 & 34.2 & 35.0 & 30.5 & 31.1 \\
\rowcolor{Gray}
$+D^l_t+D^u_{0.95}$ & 39.9 & 40.5 & 26.2 & 27.4 & 19.3 & 20.6 & 16.0 & 17.4 & 16.0 & 16.2 & 35.0 & 35.4 & 30.8 & 31.3 \\
\rowcolor{Gray}
\hspace{3mm}$+\text{Finetune}$ & 40.4 & 41.0 & 26.6 & 27.6 & 19.5 & 20.7 & 16.1 & 17.1 & 16.2 & 16.5 & 35.4 & 35.8 & 31.3 & 31.5 \\
$+D^l_t+D^u_{0.90}$ & 39.4 & 39.7 & 26.1 & 27.0 & 18.9 & 19.9 & 15.3 & 16.4 & 15.6 & 15.8 & 34.5 & 35.0 & 29.6 & 30.2 \\
\hspace{3mm}$+\text{Finetune}$ & 40.4 & 40.4 & 26.2 & 26.9 & 19.1 & 19.6 & 15.2 & 15.8 & 16.3 & 16.4 & 35.5 & 35.7 & 30.5 & 30.7 \\
$+D^l_t+D^u_{0.85}$ & 39.8 & 40.0 & 26.3 & 26.9 & 19.3 & 19.8 & 15.8 & 16.1 & 16.0 & 16.2 & 34.8 & 35.2 & 30.5 & 30.6 \\
\hspace{3mm}$+\text{Finetune}$ & 39.9 & 40.0 & 26.2 & 26.7 & 19.3 & 19.5 & 15.8 & 15.8 & 16.1 & 16.3 & 34.9 & 35.5 & 30.4 & 30.7 \\
$+D^l_t+D^u_{0.80}$ & 39.9 & 40.4 & 26.4 & 27.6 & 19.2 & 20.5 & 15.4 & 16.8 & 16.2 & 16.3 & 34.9 & 35.3 & 30.3 & 31.3 \\
\hspace{3mm}$+\text{Finetune}$ & 39.9 & 40.4 & 26.2 & 27.5 & 18.9 & 20.3 & 15.2 & 16.7 & 16.2 & 16.5 & 35.0 & 35.6 & 30.2 & 31.3 \\
$+D^l_t+D^u_{0.75}$ & 39.7 & 39.8 & 25.9 & 26.6 & 18.9 & 19.4 & 15.3 & 15.8 & 15.6 & 15.7 & 34.6 & 34.9 & 29.7 & 30.1 \\
\hspace
{3mm}$+\text{Finetune}$ & 39.8 & 39.9 & 25.9 & 26.7 & 18.8 & 19.5 & 15.3 & 15.9 & 15.7 & 15.9 & 34.7 & 35.1 & 29.6 & 30.3 \\
$+D^l_t+D^u_{0.70}$ & 40.2 & 40.5 & 26.4 & 27.2 & 19.4 & 20.1 & 15.8 & 16.4 & 16.4 & 16.5 & 35.2 & 35.5 & 30.8 & 31.0 \\
\hspace{3mm}$+\text{Finetune}$ & 40.3 & 40.6 & 26.4 & 27.1 & 19.4 & 19.9 & 15.9 & 16.0 & 16.5 & 16.6 & 35.2 & 35.7 & 30.5 & 30.9 \\
$+D^l_t+D^u_{0.50}$ & 39.3 & 39.8 & 26.2 & 27.5 & 18.9 & 20.3 & 15.2 & 16.7 & 15.7 & 16.0 & 33.9 & 34.4 & 29.4 & 30.6 \\
\hspace{3mm}$+\text{Finetune}$ & 39.5 & 40.1 & 26.3 & 27.6 & 19.0 & 20.5 & 15.4 & 17.1 & 15.8 & 16.2 & 34.2 & 34.9 & 29.3 & 30.8 \\
\midrule
\midrule
ATOMIC (full) & 42.7 & 42.9 & 29.6 & 30.0 & 22.0 & 22.5 & 18.6 & 18.7 & 29.1 & 29.7 & 51.1 & 52.7 & 74.5 & 75.4 \\
\bottomrule
\end{tabular}
\caption{
Full experiment results (\%) by GPT2 (XL) on commonsense inference generation (COMET) task. 
We evaluate the models on the original ATOMIC dev and test sets. 
$D^l_t$ stands for annotated abstract commonsense triples, and $D^u$ stands for unlabeled triples pseudo-labeled by our CAT framework.
The underfoot value is the threshold for selecting plausible pseudo labels. 
Fine-tune refers to fine-tuning back on the training set of our constructed ATOMIC subset. 
Rows with the best performance, which are reported in the paper, are colored in gray.
We also report performances by COMET trained on the complete ATOMIC training set in the bottom row.
}
\label{tab:additional_triple_tail_generation_performance}
\vspace{-0.1in}
\end{table*}

\end{document}